\documentclass[10pt]{article} %
\usepackage[preprint]{tmlr}

\usepackage{amsmath,amsfonts,bm}

\def\eqref#1{equation~\ref{#1}}

\def\1{\bm{1}}

\DeclareMathAlphabet{\mathsfit}{\encodingdefault}{\sfdefault}{m}{sl}
\SetMathAlphabet{\mathsfit}{bold}{\encodingdefault}{\sfdefault}{bx}{n}

\usepackage{hyperref}
\usepackage{url}

\usepackage{xcolor}
\usepackage{subfig}
\usepackage{booktabs}
\usepackage{balance}
\usepackage{graphicx}

\title{MetisFL: An Embarrassingly Parallelized Controller for \\ Scalable \& Efficient Federated Learning Workflows}

\author{\name Dimitris Stripelis \email stripeli@isi.edu \\
      \addr Information Science Institute\\
      University of Southern California
      \AND
      \name Chrysovalantis Anastasiou \email canastas@usc.edu \\
      \addr Viterbi School of Engineering\\ 
      University of Southern California
      \AND
      \name Patrick Toral \email pjtoral@isi.edu\\
      \addr Information Science Institute\\
      University of Southern California
      \AND
      \name Armaghan Asghar \email asghar@usc.edu\\
      \addr Viterbi School of Engineering\\ 
      University of Southern California
      \AND      
      \name Jos\'{e} Luis Ambite \email ambite@isi.edu\\
      \addr Information Science Institute\\
      University of Southern California
}

\def\month{05}  %
\def\year{2023} %

\begin{document}

\maketitle

\begin{abstract}
A Federated Learning (FL) system typically consists of two core processing entities: the federation controller and the learners. The controller is responsible for managing the execution of FL workflows across learners and the learners for training and evaluating federated models over their private datasets. While executing an FL workflow, the FL system has no control over the computational resources or data of the participating learners. Still, it is responsible for other operations, such as model aggregation, task dispatching, and scheduling. These computationally heavy operations generally need to be handled by the federation controller. Even though many FL systems have been recently proposed to facilitate the development of FL workflows, most of these systems overlook the scalability of the controller. To meet this need, we designed and developed a novel FL system called MetisFL, where the federation controller is the first-class citizen. MetisFL re-engineers all the operations conducted by the federation controller to accelerate the training of large-scale FL workflows. By quantitatively comparing MetisFL against other state-of-the-art FL systems, we empirically demonstrate that MetisFL leads to a 10-fold wall-clock time execution boost across a wide range of challenging FL workflows with increasing model sizes and federation sites.
\end{abstract}

\maketitle

\section{Introduction}

\begin{figure*}[htpb]
  \centering  
  \includegraphics[width=\linewidth]{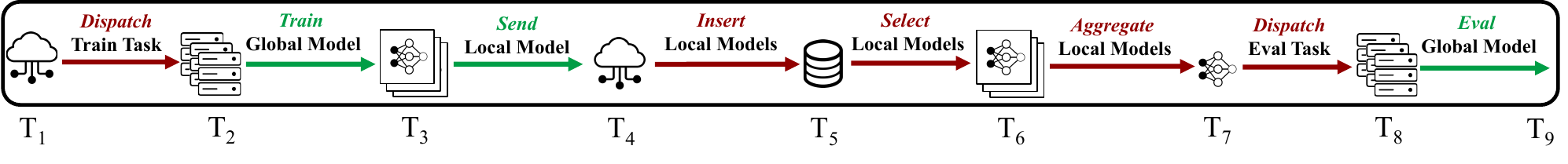}
  \captionsetup{justification=centering}
  \caption{A typical Federated Learning workflow. The red color represents operations executed by the controller. The green color represents operations executed by the learner(s).}
  \label{fig:FederatedWorkflow}
\end{figure*}

Federated Learning (FL) has emerged as a standard distributed learning approach for training machine and deep learning models across dispersed data sources that cannot share data due to regulatory or privacy concerns. During the execution of an FL workflow, data always remain at the source, and sources only share their locally trained model parameters. Even though this approach mitigates the problem of direct data leakage, it creates many new system challenges related to the design, development, and deployment of FL workflows. Recently, several open-source FL frameworks have become available to meet this need, including but not limited to Nvidia FLARE~\cite{roth2022nvidia}, Flower~\cite{beutel2022flower}, FedML~\cite{he2020fedml}, IBM FL~\cite{ibmfl2020ibm}, OpenFL~\cite{reina2021openfl}, LEAF~\cite{caldas2018leaf}, and TensorFlow Federated~\cite{googleTFF} (TFF). Some systems (e.g., Flower, TFF) make it easier to adapt the execution of centralized ML models into federated settings and implement new FL algorithms, while others (e.g., NVFlare, OpenFL, IBM FL) facilitate the transition of existing ML/DL model training pipelines into production environments.

Even though each system offers unique solutions and optimizations, one processing entity is often overlooked: the federation controller/aggregator. Typically, a centralized FL environment~\footnote{In our work, we only focus on centralized environments, even though other topologies exist as well~\cite{rieke2020future}, such as decentralized and hierarchical.} consists of a single controller and a set of participating clients/learners. Even though the controller has no ownership over the computational resources or the data of the participating learners in the federation, it is still responsible for managing the execution of the FL workflows across learners, scheduling and dispatching the local training tasks, and receiving, storing, and aggregating learners' model parameters. When conducting large-scale FL workflows with hundreds of learners and very large models, the controller must efficiently manage all available resources and perform all required operations with the minimum cost. On these grounds, we consider the controller the primary bottleneck of any FL system when handling the orchestration of large-scale workflows.

In Figure~\ref{fig:FederatedWorkflow}, we present the sequence of these operations for a typical FL workflow. At Timestamp T1, the controller receives an update request from a participating learner(s). Timestamps T2-T4 represent learner(s)' local model training and local model parameter transfer, and timestamps T4-T7 represent the reception, storing, selection, and aggregation of learner(s)' locally trained model parameters by the controller. Finally, timestamps T7-T9 represent the scheduling and dispatching of the newly computed federated model to the learners for evaluation. All operations occurring between timestamps T1-T9 represent the federation round or community update request for synchronous and asynchronous settings~\cite{stripelis2022semi}, respectively. Timestamps T1-T4 represent the training round for a single (asynchronous) or multiple learners (synchronous), timestamps T4-T7 local model consolidation and aggregation, and timestamps T7-T9 the evaluation round for a single (asynchronous) or multiple learners (synchronous).

To this end, we designed and developed from scratch a novel FL system, called~\textit{MetisFL}, where the federation controller is the first-class citizen of the system. To boost the performance of the controller's operations, we re-engineered the controller in C++ and optimized weight tensor processing and network transmission. MetisFL achieves a 10-fold to 100-fold wall-clock time execution improvement against other leading FL systems. MetisFL is the first FL system that accelerates the training of FL workflow by optimizing the controller's operations. Our study is the first to stress-test these operations across different FL systems.

\section{Federated Learning Systems}
Various FL systems have been recently introduced~\cite{he2020fedml,beutel2022flower,caldas2018leaf,googleTFF,roth2022nvidia,reina2021openfl,ibmfl2020ibm}, with each one aiming to optimize different aspects of a particular FL workload. 
FedML~\cite{he2020fedml} is an FL framework that supports standalone simulation, distributed training, cross-silo, and cross-device FL applications. FedML uses PyTorch as its core training engine and supports communication protocols like NCCL, MPI, and MQTT. Flower~\cite{beutel2022flower} is an open-source FL framework designed to be an extensible and scalable federated ML/DL model agnostic framework targeted for heterogeneous client environments for simulation and real-world experiments. It supports a wide range of machine and deep learning computing engines (e.g., TensorFlow, PyTorch, JAX) and uses gRPC for communication between participating clients. NVIDIA Flare (NVFlare)~\cite{roth2022nvidia}, OpenFL~\cite{reina2021openfl}, and IBM FL~\cite{ibmfl2020ibm} were designed primarily for production-oriented environments. NVFlare supports a high-availability controller with fail-over capabilities and provides a domain-agnostic FL SDK for implementing a wide range of FL applications. OpenFL~\cite{reina2021openfl} originated from the collaboration of Intel Labs and the University of Pennsylvania as part of a research project for FL applications in healthcare. The framework has now become domain-agnostic, supporting various use cases. At its core, OpenFL defines an FL plan with all the configurations necessary to execute an FL workload across learners. IBM FL's key design is its fast start-up time for enterprise applications. IBM FL aims to minimize the learning curve for a data scientist to migrate existing centralized machine learning workloads into federated settings by enabling the design of custom fusion (aggregation) algorithms and providing easy deployment of workloads across computing environments.

When it comes to systematically comparing the various FL systems, previous studies have focused on the qualitative aspects of existing FL systems and have taxonomized them based on functional capabilities and architectural designs~\cite{li2021survey,liu2022distributed,kairouz2021advances}. Other studies~\cite{liu2022unifed,hu2022oarf} have focused on quantitatively benchmarking these systems by executing FL workloads that include challenging ML/DL model architectures and non-IID dataset distributions but whose aim was to measure the communication and computational overhead required to reach particular FL model learning performances. In our work, we take inspiration from these studies and perform a qualitative comparison between MetisFL and other leading frameworks, as shown in Table~\ref{tbl:FLSystems_Comparison}. However, our quantitative analysis does not focus on traditional measures of the FL model's performance (e.g., model accuracy, communication, and computation cost). Instead, we conduct an end-to-end stress test analysis across frameworks to quantify the system scalability occurring from the operations performed at the controller, which is often a major scalability bottleneck in FL systems. 
\section{MetisFL: Design \& Architecture}

\begin{figure*}[htpb]
    \centering
    \includegraphics[width=\linewidth]{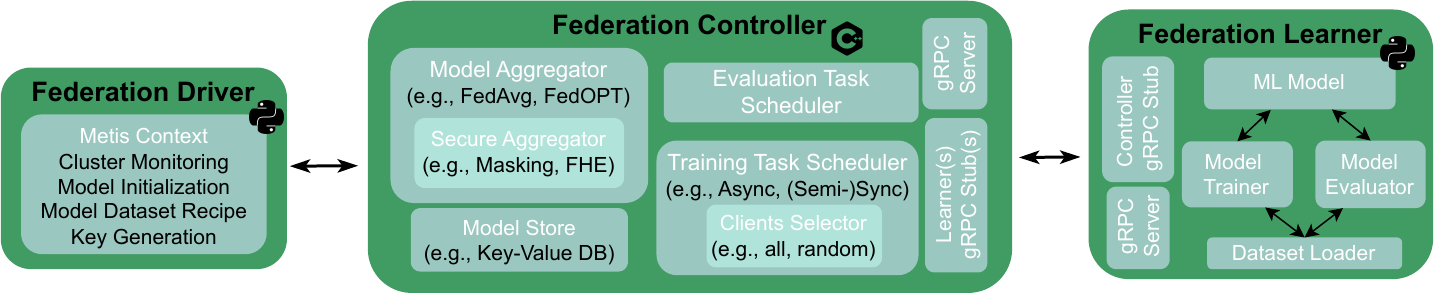}
    \captionsetup{justification=centering}
    \caption{MetisFL Internal Components.}
    \label{fig:MetisFL_Components_Details}
\end{figure*}

MetisFL~\footnote{\url{https://github.com/bioint/MetisFL}} is designed from the bottom up to comply with the architectural principles of modularity (implementation of micro-services), extensibility (expanding system functionalities), and configurability (ease of use)~\cite{beutel2022flower}. MetisFL is domain agnostic and can be used to conduct various FL workflows across organizations (a.k.a. cross-silo) or devices (a.k.a. cross-device)~\cite{yang2019federated}. Figure~\ref{fig:MetisFL_Components_Details} shows a detailed overview of the MetisFL internal architecture. The framework consists of three core components, the \textit{Federation Driver}, the \textit{Federation Controller}, and the \textit{Federation Learner}, similar to the programming model of the Apache Spark framework~\cite{zaharia2010spark}.

\begin{figure}[htpb]
    \centering
    \includegraphics[width=0.99\linewidth]{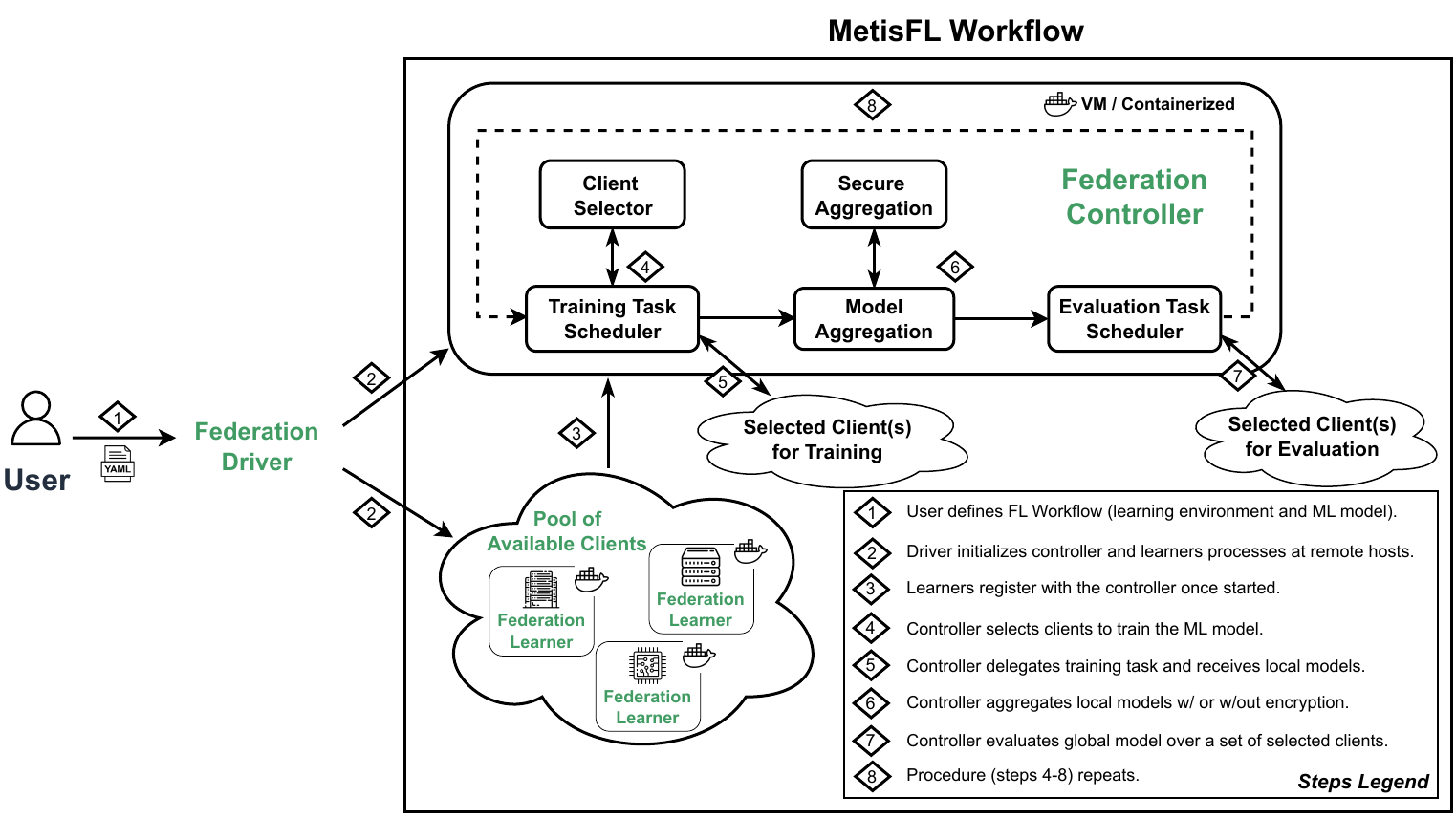}
    \captionsetup{justification=centering}
    \caption{MetisFL Workflow.}
    \label{fig:MetisFL_FLWorkflow}
\end{figure}

As shown in Figure~\ref{fig:MetisFL_FLWorkflow}, when a system user needs to run an FL workload, he/she needs to wrap the original Keras/PyTorch model architecture around the MetisFL model abstraction and materializing specific functions that the system can use to run the model operations (e.g., fit and evaluate functions for PyTorch models). Subsequently, the user needs to specify how the data will be loaded in the model (data recipe) and the federated environment (yaml file) that contains all the model (e.g., optimizer) and running environment (e.g., host machines) configurations. Once all this information is defined, the Federation Driver parses the FL workflow and creates the MetisFL Context. The MetisFL Context is responsible for initializing and monitoring the federation cluster, initializing the original model state, shipping the model state to each learner, defining the data loading recipe for each learner, and generating the security keys where needed, e.g., SSL certificates (see also Figure~\ref{fig:MetisFL_SSL} in Appendix) and/or FHE key pair~\cite{stripelis2021secure,stripelis2022secure}. When all required services are alive, the Federation Controller manages the federation cluster by selecting and delegating training and evaluating tasks that need to be performed by the Federation Learners (cluster nodes) over their private datasets and storing and aggregating the learners' local models (w/ or w/out encryption) when local training is complete. 

The MetisFL prototype was initially proposed in the work of~\cite{stripelis2022semi}. However, since then, substantial improvements have been made in code execution and parallelism. In its original implementation, the MetisFL controller was developed in Python. However, due to Python's limited memory management capabilities, the Python version led to a high latency when aggregating large-sized models and/or aggregating models from a large pool of participating clients (e.g., $>$ 100). Moreover, in its original implementation, as more clients were considered in the federation and the size of the ML/DL models became larger, the training and evaluation tasks scheduling and dispatching became extremely slow. Given that Python relies on the Global Interpreter Lock (GIL) for proper thread management, the concurrent execution of tasks was not possible, dramatically slowing down the overall execution time of an FL workflow. For these reasons, the MetisFL architecture was redesigned and the controller was refactored in a native C++ implementation.

\begin{figure}[htpb]
    \centering
    \includegraphics[width=0.5\linewidth]{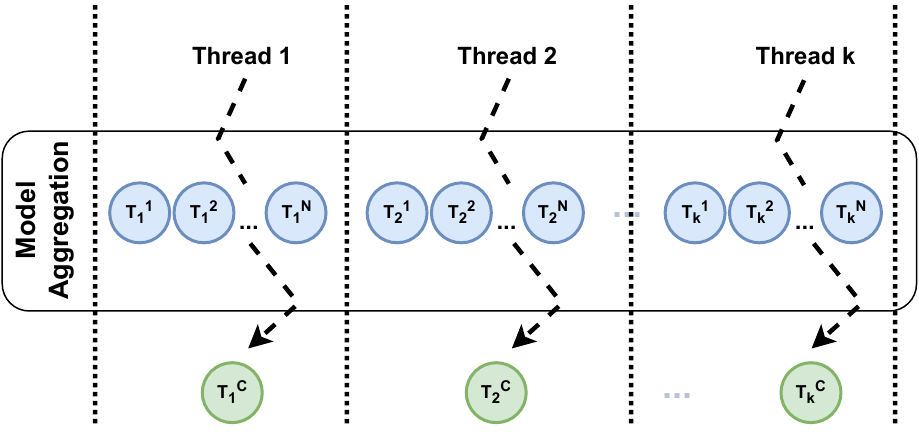}
    \captionsetup{justification=centering}
    \caption{MetisFL Parallelized Model Aggregation.}
    \label{fig:MetisFL_Aggregation_WithOpenMP}
\end{figure}

In terms of redesign, compared to other systems (e.g., Flower, FedML), in MetisFL, the ML/DL model is transferred over the network as a sequence of tensors with each tensor being represented in a byte protobuf data type. This allows different gRPC services (i.e., controller-learners) to communicate with each other with very low overhead. This functionality is accomplished by first flattening each tensor/matrix of the ML/DL model, then dumping the tensor (as bytes), and finally constructing a proto message that represents the structure of the original tensor to enable reconstruction after proto tensor reception e.g., tensor's byte order and data type. 

With respect to the controller's process refactoring, the controller can now leverage the underlying system's hardware capabilities and parallelize most of its computationally heavy operations, i.e., task scheduling and dispatching and model aggregation. Especially in the case of model aggregation, by taking advantage of the tensor-based model representation, the controller parallelizes the aggregation of multiple model tensors over all available cores to speed up the operation. As shown in Figure~\ref{fig:MetisFL_Aggregation_WithOpenMP}, for a federation of N learners and a model with k tensors, MetisFL uses one thread per model tensor to compute the aggregated tensor. For instance, to calculate the aggregated tensor for the first model tensor ($T_1^C$), a single thread is used to aggregate the N learners' tensors (i.e., tensor $T_1^1 ... T_1^N$); thread parallelism is enabled using OpenMP~\cite{dagum1998openmp}.

\begin{table*}[htbp]
\scriptsize
\begin{tabular}{@{}lcccccc@{}}

\toprule
\textbf{Dimension} & Nvidia FLARE & Flower & FedML & IBM FL & OpenFL & MetisFL*  \\
\cmidrule(){1-1}
\cmidrule(lr){2-2}
\cmidrule(lr){3-3}
\cmidrule(lr){4-4}
\cmidrule(lr){5-5}
\cmidrule(lr){6-6}
\cmidrule(lr){7-7}
\textbf{Deployment} & & & & & & \\
\quad Standalone & \checkmark & \checkmark & \checkmark & \checkmark & \checkmark & \checkmark \\
\quad Distributed & \checkmark & \checkmark & \checkmark & \checkmark & \checkmark & \checkmark \\
\quad Cross-Silo & \checkmark & \checkmark & \checkmark & \checkmark & \checkmark & \checkmark \\ 
\quad Cross-Device & $\times$ & \checkmark & \checkmark & $\times$ & $\times$ & \checkmark \\
\quad Containerized & \checkmark & \checkmark & \checkmark & \checkmark & $\times$ & \checkmark \\
\textbf{ML Environment} & & & & & & \\
\quad Model Types & ML$\mid$DL & ML$\mid$DL & ML$\mid$DL & ML$\mid$DL & ML$\mid$DL & ML$\mid$DL \\ 
\quad Backend & Torch$\mid$TF$\mid$MONAI & Torch$\mid$TF$\mid$MX$\mid$JAX & Torch$\mid$TF$\mid$MX$\mid$JAX & Torch$\mid$TF & Torch$\mid$TF & Torch$\mid$TF \\ 
\quad LocalOpt & \checkmark & \checkmark & \checkmark & \checkmark & \checkmark & \checkmark \\ 
\quad GlobalOpt & \checkmark & \checkmark & \checkmark & \checkmark & \checkmark & \checkmark \\ 
\textbf{Data Partitioning} & & & & & & \\
\quad Horizontal & \checkmark & \checkmark & \checkmark & \checkmark & \checkmark & \checkmark \\ 
\quad Vertical & $\times$ & $\times$ & $\times$ & $\times$ & $\times$ & $\times$ \\ 
\textbf{Privacy \& Security} & & & & & & \\
\quad Private Training & \checkmark & \checkmark & \checkmark & \checkmark & \checkmark & \checkmark \\ 
\quad Secure Aggregation & FHE & Masking$\mid$FHE & Masking$\mid$FHE & FHE & TEE & FHE \\
\quad Crypto Library & TenSeal & native & native & HElayers & Graphene & PALISADE \\ 
\textbf{Communication} & & & & & & \\
\quad Centralized & \checkmark & \checkmark & \checkmark & \checkmark & \checkmark & \checkmark \\ 
\quad Decentralized & $\times$ & $\times$ & $\checkmark$ & $\times$ & $\times$ & $\times$ \\ 
\quad Hierarchical & $\times$ & $\times$ & $\times$ & $\times$ & $\times$ & $\times$ \\ 
\quad TLS & \checkmark & \checkmark & \checkmark & \checkmark & \checkmark & \checkmark \\ 
\quad Network & gRPC & gRPC & MPI & AMQP & gRPC &  gRPC \\
\textbf{Communication Protocol} & & & & & & \\
\quad Synchronous & \checkmark & \checkmark & \checkmark & \checkmark & \checkmark & \checkmark \\ 
\quad Asynchronous & $\times$ & $\times$ & $\times$ & $\times$ & $\times$ & \checkmark \\ 
\textbf{Software} & & & & & & \\ 
\quad End-user & Python & Python & Python & Python & Python & Python \\ 
\quad Learner & Python & Python & Python & Python & Python & Python \\
\quad Aggregator & Python & Python & Python & Python & Python & C++ \\ 

\bottomrule

\end{tabular}
\captionsetup{justification=centering}
\caption{A qualitative comparison of different Federated Learning Systems.}
\label{tbl:FLSystems_Comparison}
\end{table*}

\section{Evaluation}
We perform a qualitative and quantitative analysis to compare the trade-offs across leading FL frameworks~\footnote{We do not compare against LEAF~\cite{caldas2018leaf} and TFF~\cite{googleTFF} frameworks since both frameworks are used primarily to experiment with new FL optimization algorithms~\cite{li2021survey}.}. The aim of our evaluation is two-fold. Through the qualitative comparison, we aim to understand the out-of-the-box support provided by the different FL frameworks, and through the quantitative evaluation, we aim to gain insights into how different implementations of the federation controller can affect the overall execution of an FL workflow. Following the taxonomy proposed in the works of~\cite{li2021survey, kairouz2021advances,liu2022unifed}, we report our qualitative comparison in Table~\ref{tbl:FLSystems_Comparison}, while Figures~\ref{fig:FL_Systems_Quantitative_Comparison_100k},~\ref{fig:FL_Systems_Quantitative_Comparison_1M}, and~\ref{fig:FL_Systems_Quantitative_Comparison_10M} present the execution time of the various operations, in isolation, occurring during a typical FL workflow (see Figure~\ref{fig:FederatedWorkflow}). It is critical to note that in the current evaluation, we assume that all local models fit in the controller's in-memory store (e.g., hash map). Therefore, we consider model insertion and selection operations to take constant time across all frameworks (see section~\ref{sec:Discussion}).

\subsection{Qualitative Evaluation}
We compare leading FL systems across several categories.

The~\textit{Deployment} category refers to the case where a federation can be run in a simulated environment as parallel processes within a single server or distributed across multiple nodes (servers). All systems support standalone and distributed execution and deployment for cross-silo settings. However, OpenFL, Nvidia FLARE, and IBM FL do not support execution in cross-device settings, with OpenFL having no support for containerized execution. 

In the~\textit{ML Environment} category, model types describe the machine learning models each system can support, backend the machine/deep learning engine used to perform model training, evaluation, and inference, and LocalOpt and GlobalOpt whether the system allows the development of customized local (learner) and global (controller) function optimization algorithms. In this category, all frameworks support the execution of various model architectures and both the PyTorch and Tensorflow (TF) backend engines. Our system (MetisFL) currently only supports PyTorch and Tensorflow; however, it can be easily extended to support other training engines, e.g., MONAI, MXNet(MX), JAX.

In the case of~\textit{Data Partitioning}, we evaluate whether each system can support learning over different partitioning schemes. All systems readily support horizontally partitioned learning environments. In their documentation, FedML also states that their platform can be extended to support vertical learning scenarios; however, support is not provided out of the box.

For the~\textit{Privacy \& Security} category, we assess whether a system can support private learning (differential privacy), the cryptographic method type used to perform secure aggregation operations, and which library is used for the cryptography operations. All systems support differential private learning. Nvidia FLARE, IBM FL, and MetisFL support homomorphic operations through the CKKS~\cite{ckks_paper} scheme. For FHE operations, NVFlare uses the TenSeal library~\cite{seal}, IBM FL the HElayers library~\cite{aharoni2020helayers}, and MetisFL the PALISADE library~\cite{palisade}. OpenFL operates on a hardware-integrated trusted execution environment (TEE~\cite{sabt2015trusted}), while Flower (Salvia~\cite{li2021secure}) and FedML (LightSecAgg~\cite{so2022lightsecagg}) both utilize a mask-based encryption approach; both frameworks recently introduced the support for FHE protocols. Concerning the cryptography library, all systems (including ours) depend on an external library, while Flower and FedML provide native implementations for the required masking operations.

In the \textit{Communication} category, we compare the federated learning topologies under which each system can operate, whether the communication across all participating parties is performed within an encrypted channel (TLS), and what network protocol is used to exchange messages. All systems can operate in a centralized federated learning environment (one aggregator,  multiple clients). FedML provides support for decentralized settings (peer-to-peer), with FedML stating in their documentation support for hierarchical federated settings. Finally, every system uses gRPC to establish communication across all federation parties, except for FedML and IBM FL, which use the MPI and AMQP protocols.

Another category in which we observe limited implementation capabilities across existing systems is the~\textit{Communication Protocol}. Even though systems provide support for synchronous communication and aggregation, they lack support for asynchronous protocols. In contrast, MetisFL provides both synchronous (including semi-synchronous~\cite{stripelis2022semi}) and asynchronous execution. Based on other systems' documentation, Flower and FedML are the only systems planning to support asynchronous execution~\cite{stripelis2022secure}. 

Finally, compared to previously proposed metrics~\cite{li2021survey, kairouz2021advances,liu2022unifed}, we also consider the programming language used to develop each component in the federated learning system: aggregator, learner and end-user API. All three components for all presented systems are developed in Python. However, in our framework, the aggregator is developed in C++.

\subsection{Quantitative Evaluation}

\begin{figure*}

    \centering
    \subfloat[Train Task Dispatch]{
        \centering
        \includegraphics[width=0.25\linewidth]{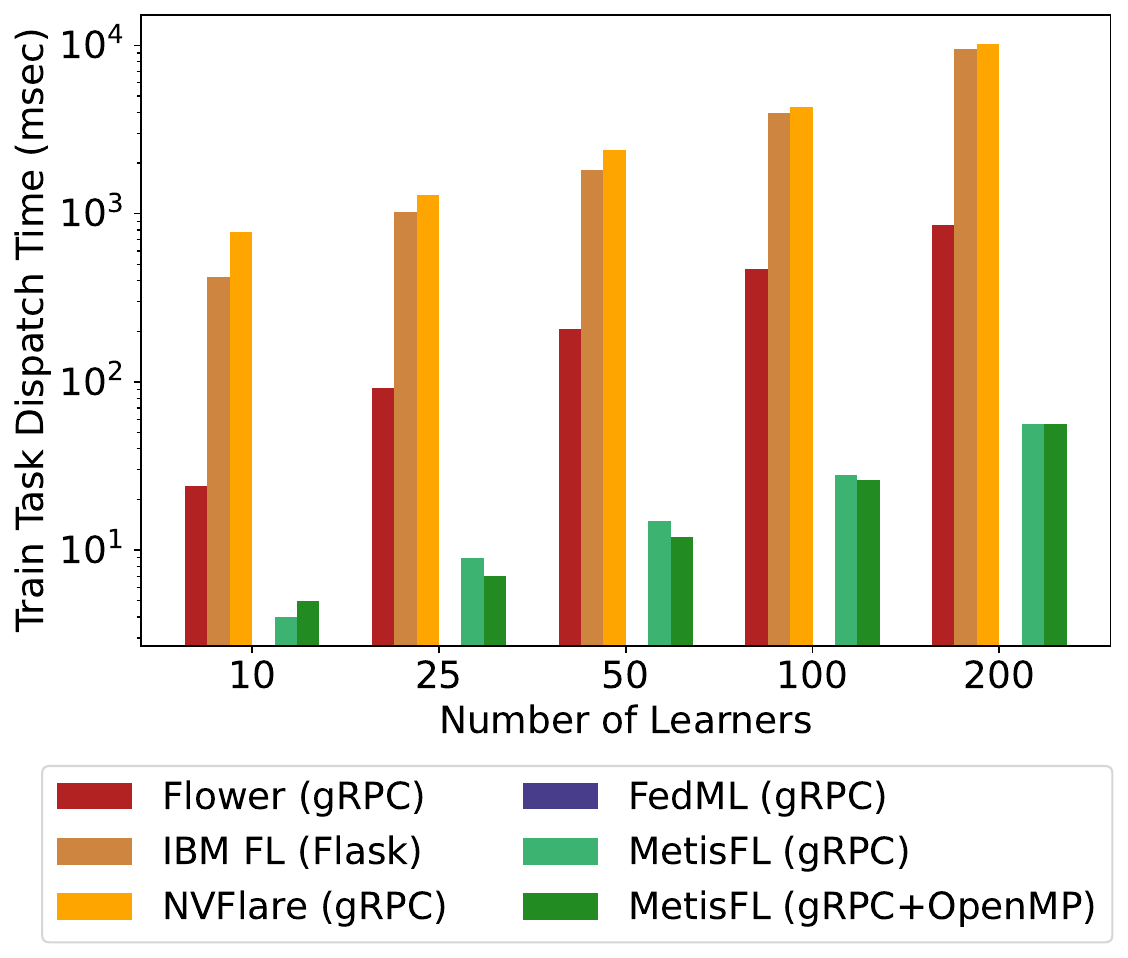}
        \label{fig:params_100k_train_dispatch_time}
    }
    \subfloat[Train Round]{
        \centering
        \includegraphics[width=0.25\linewidth]{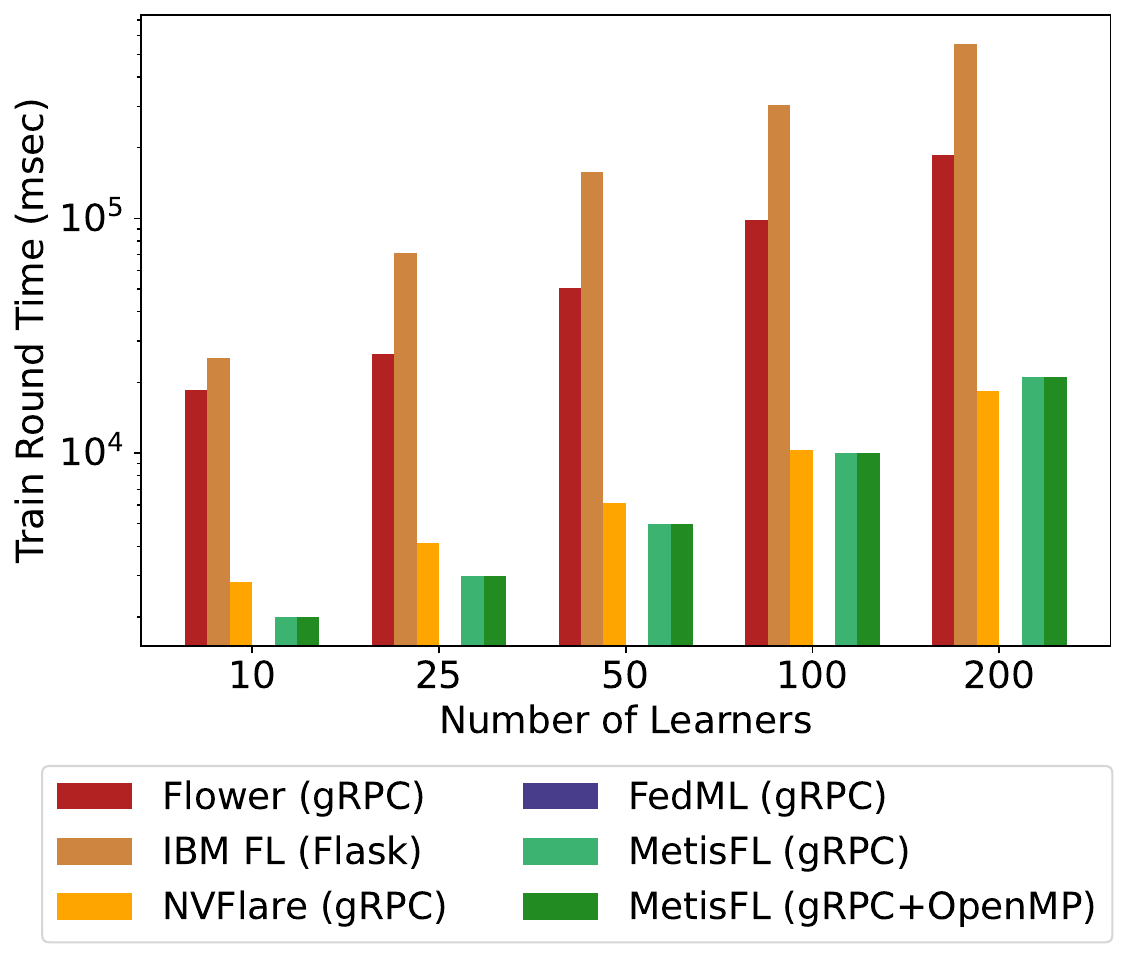}
        \label{fig:params_100k_train_round_time}
    }    
    \subfloat[Aggregation Time]{
        \centering
        \includegraphics[width=0.25\linewidth]{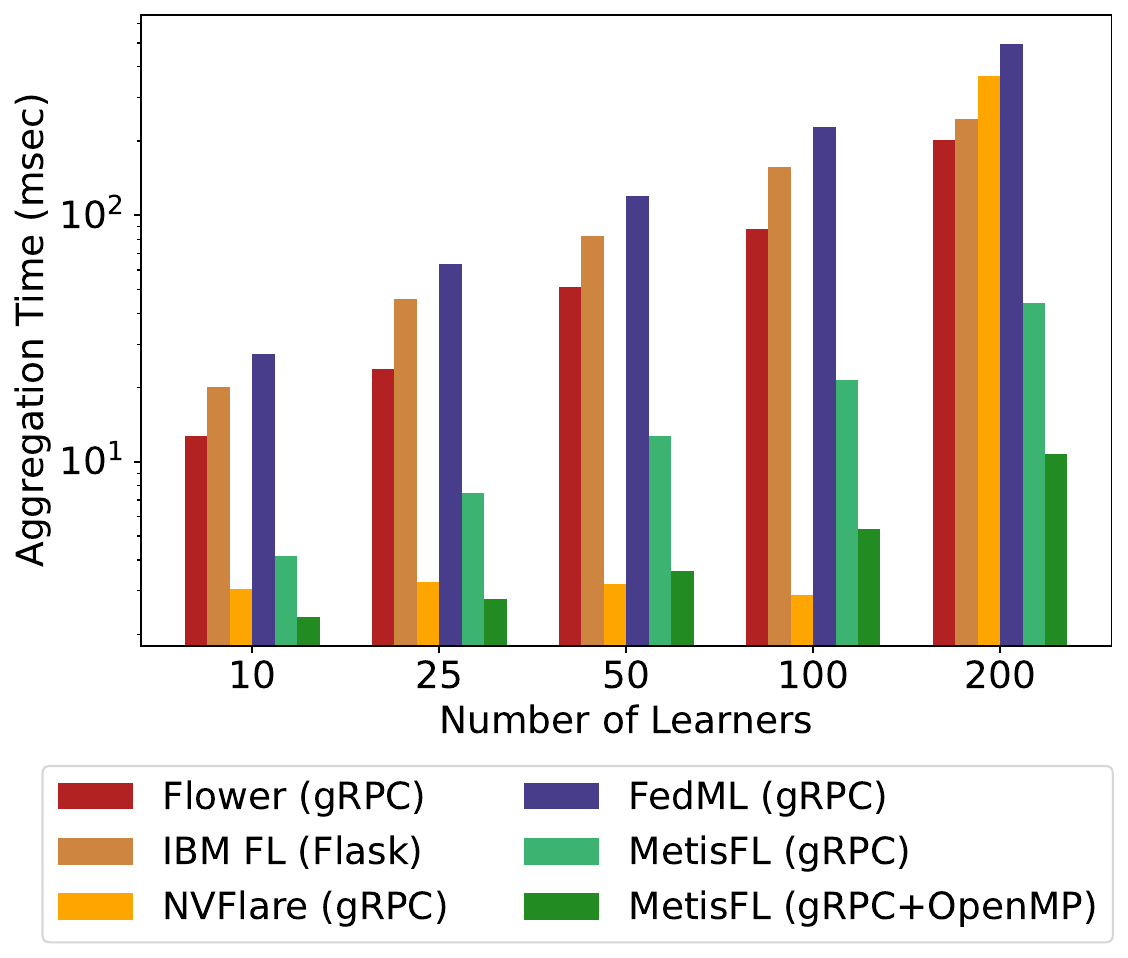}
        \label{fig:params_100k_agg_time}
    }
    
    \subfloat[Eval Task Dispatch]{
        \centering
        \includegraphics[width=0.25\linewidth]{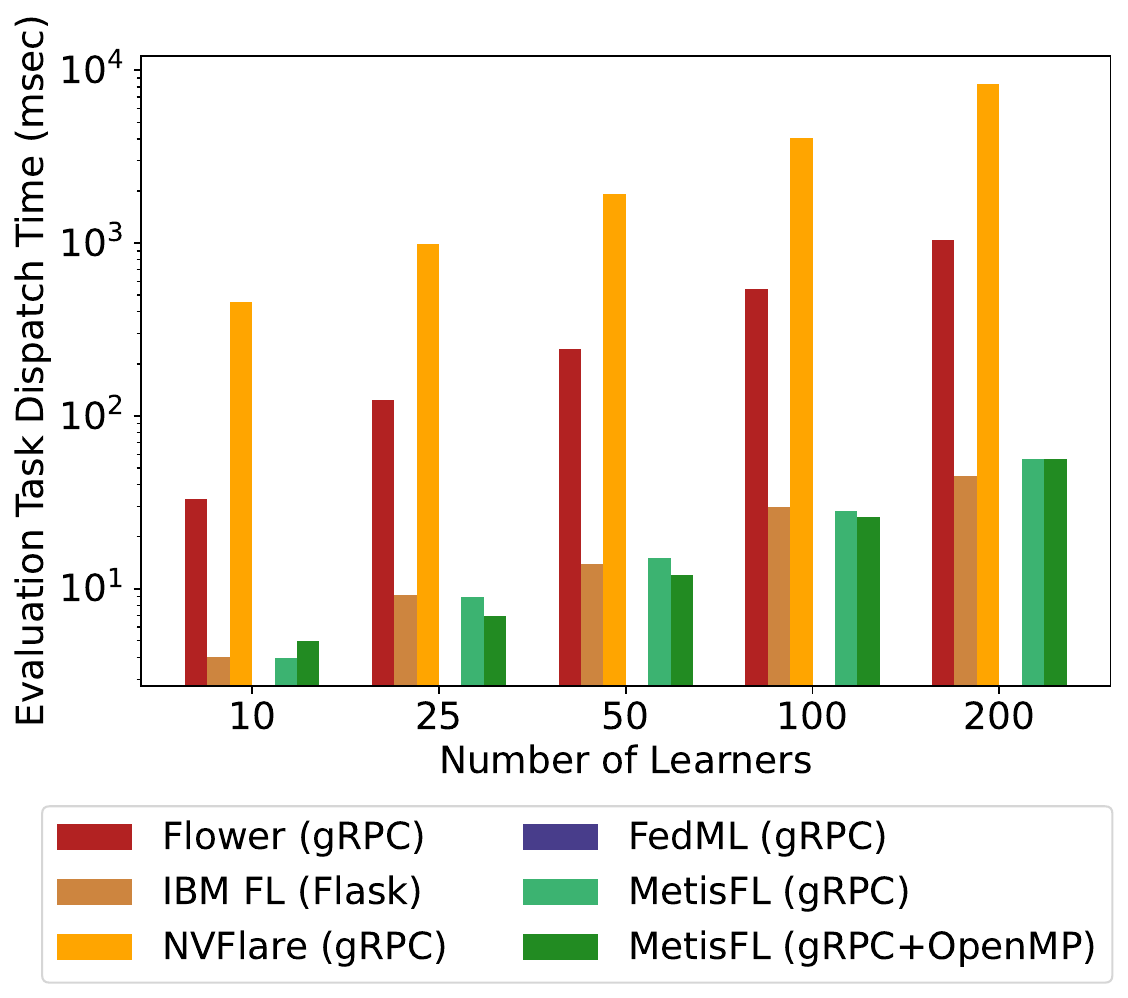}
        \label{fig:params_100k_eval_dispatch_time}
    }
    \subfloat[Eval Round]{
        \centering
        \includegraphics[width=0.25\linewidth]{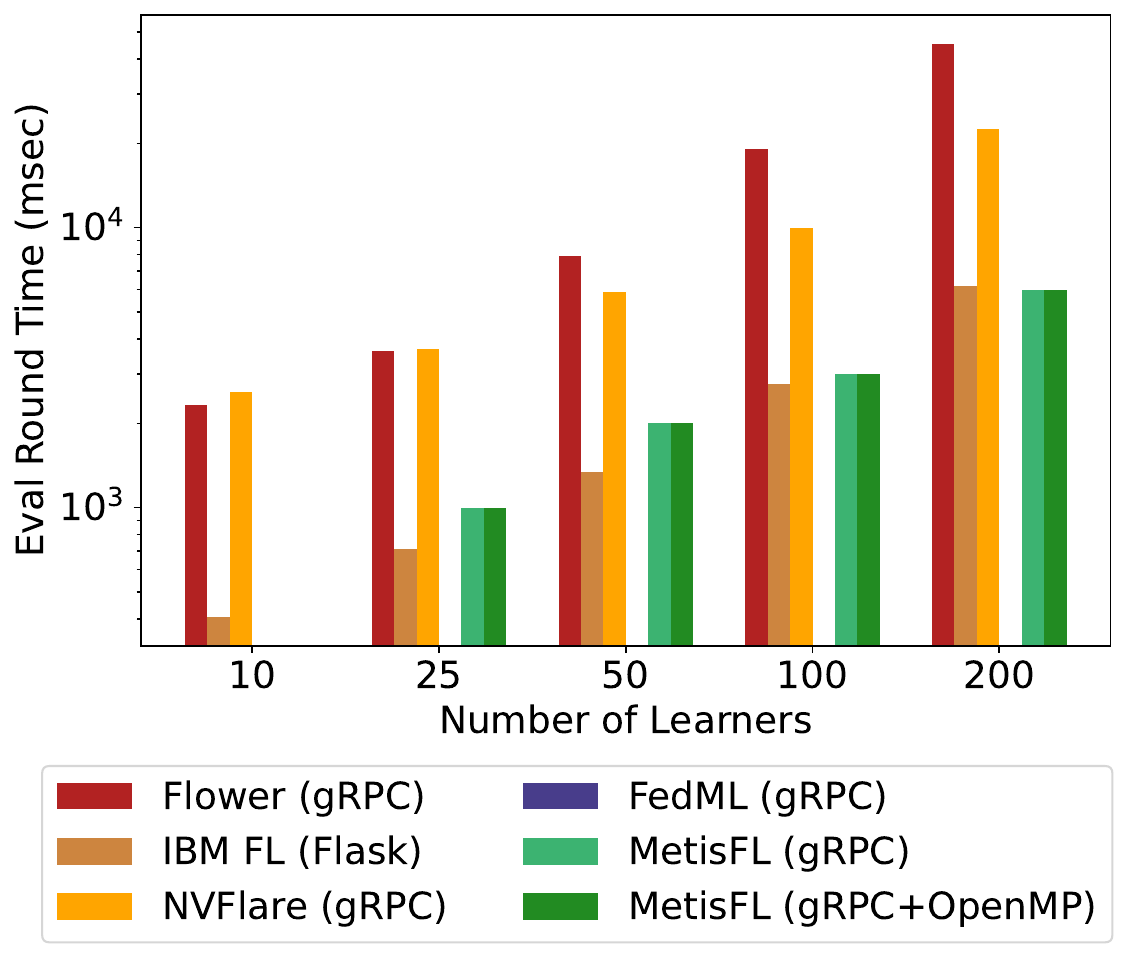}
        \label{fig:params_100k_eval_round_time}
    }
    \subfloat[Federation Round]{
        \centering
        \includegraphics[width=0.25\linewidth]{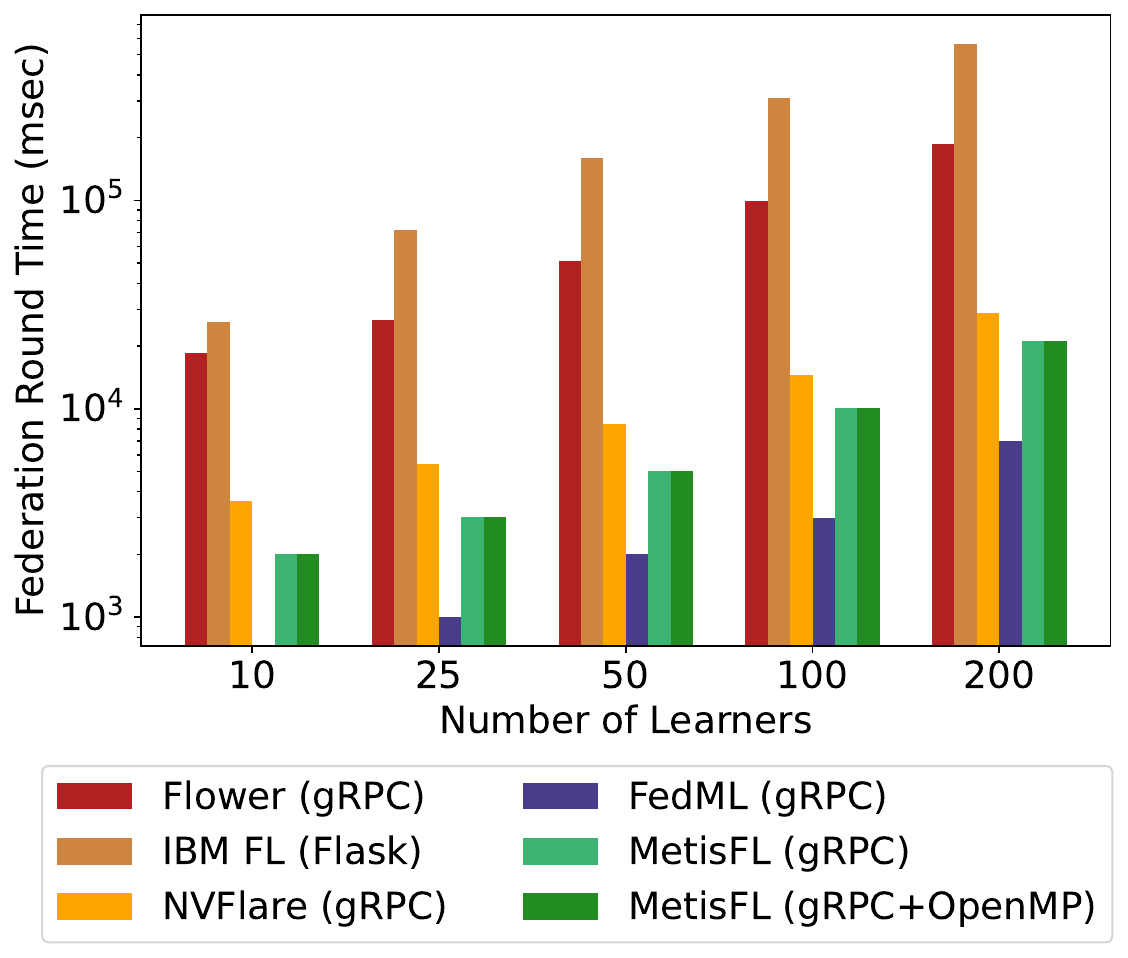}
        \label{fig:params_100k_federation_round_time}
    }        

    \captionsetup{justification=centering}
    \caption{FL frameworks operations comparison for 100k parameters (the y-axis is in logscale).}
    \label{fig:FL_Systems_Quantitative_Comparison_100k}
\end{figure*}

\begin{figure*}

    \centering
    \subfloat[Train Task Dispatch]{
        \centering
        \includegraphics[width=0.25\linewidth]{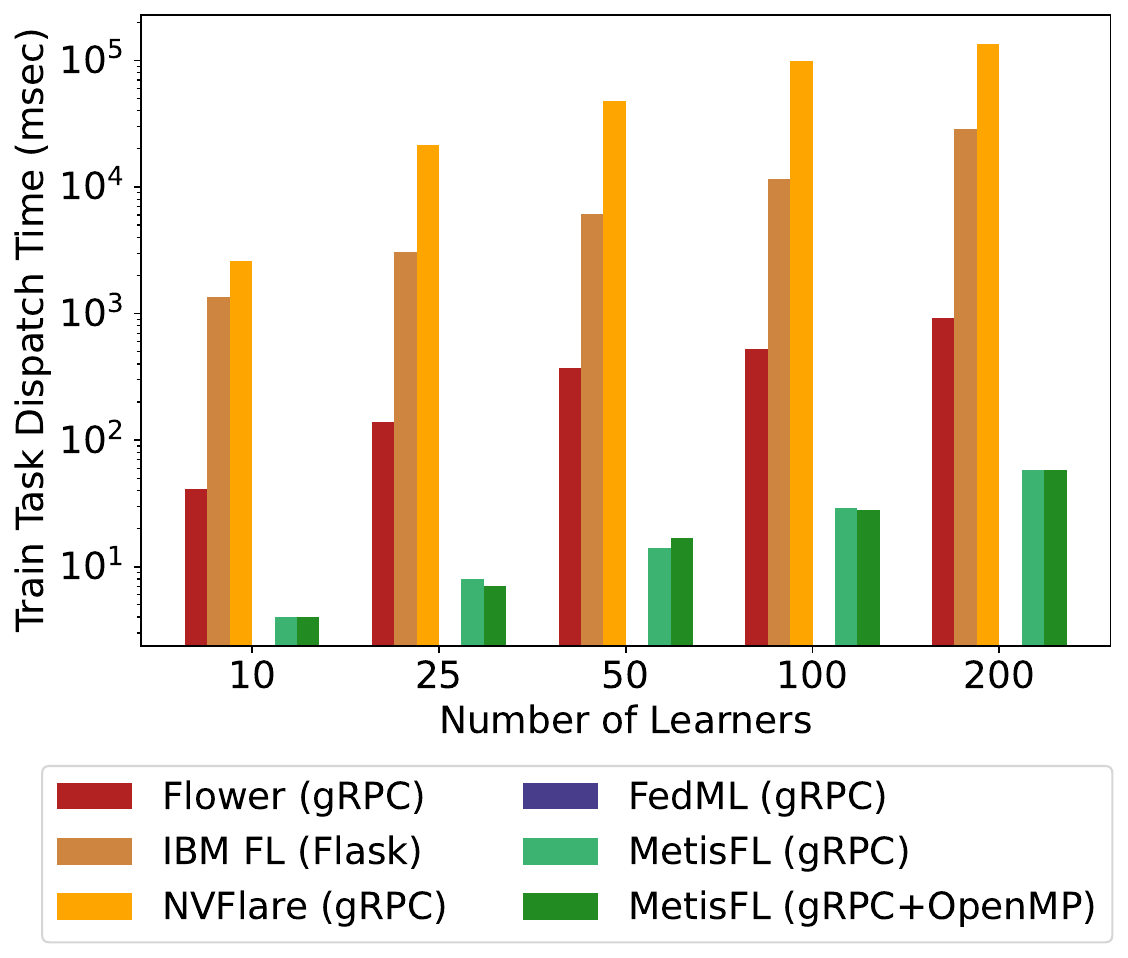}
        \label{fig:params_1M_train_dispatch_time}
    }
    \subfloat[Train Round]{
        \centering
        \includegraphics[width=0.25\linewidth]{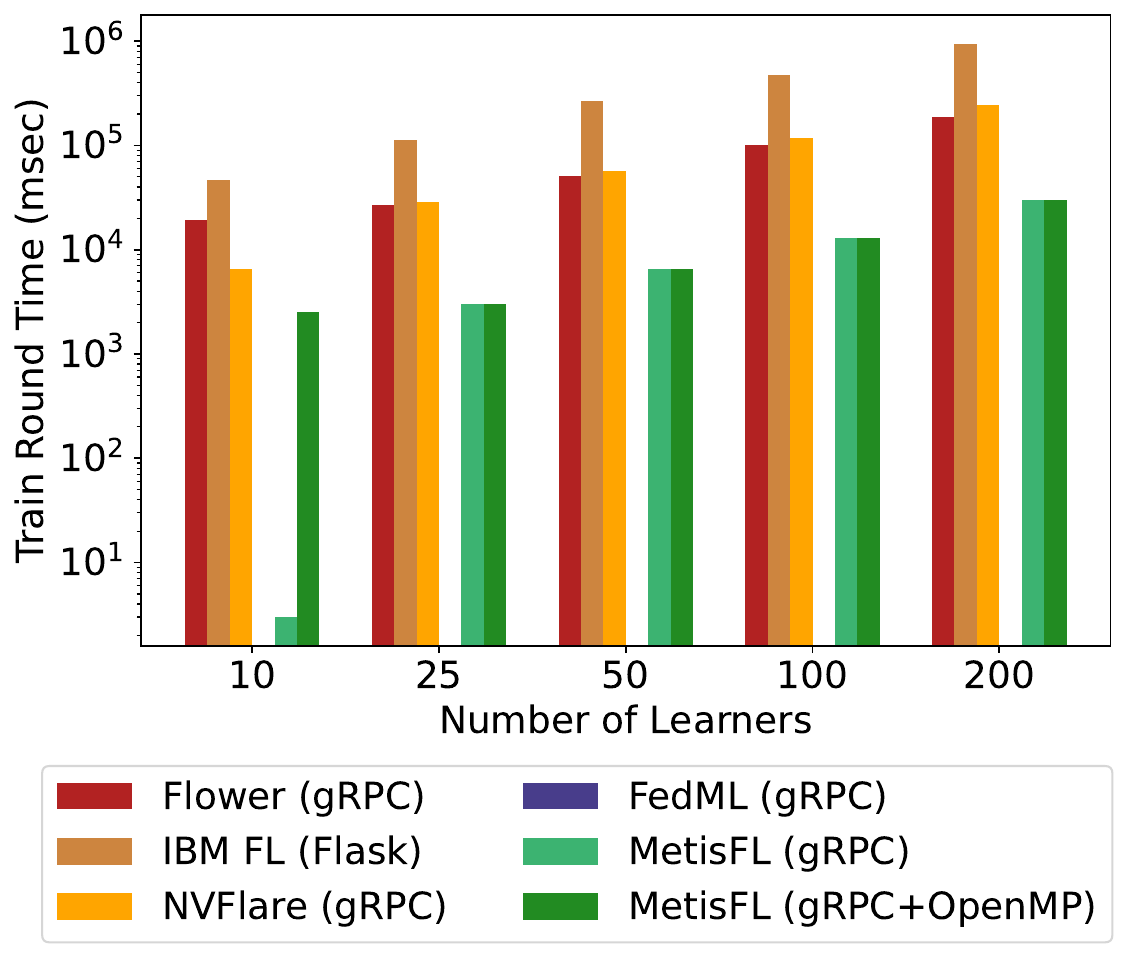}
        \label{fig:params_1M_train_round_time}
    }    
    \subfloat[Aggregation Time]{
        \centering
        \includegraphics[width=0.25\linewidth]{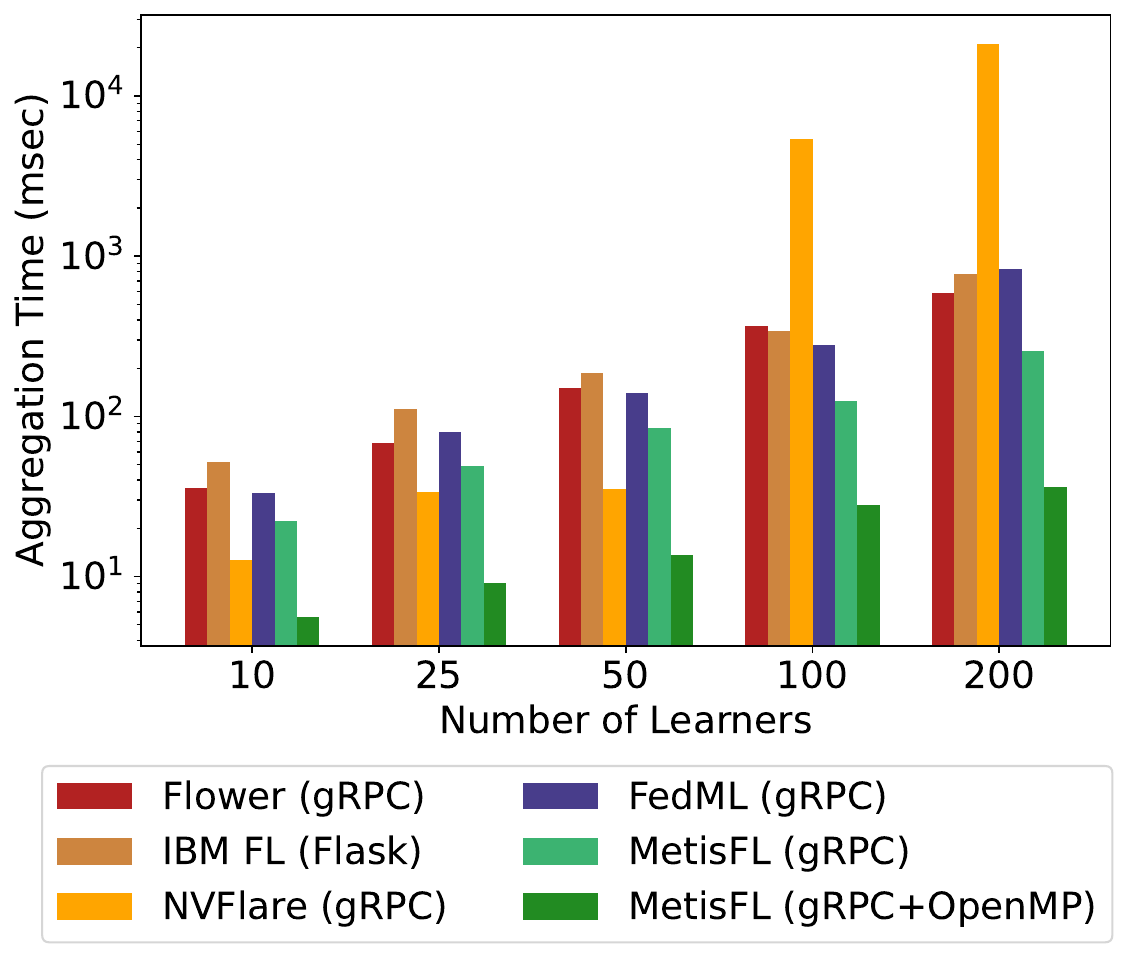}
        \label{fig:params_1M_agg_time}
    }
    
    \subfloat[Eval Task Dispatch]{
        \centering
        \includegraphics[width=0.25\linewidth]{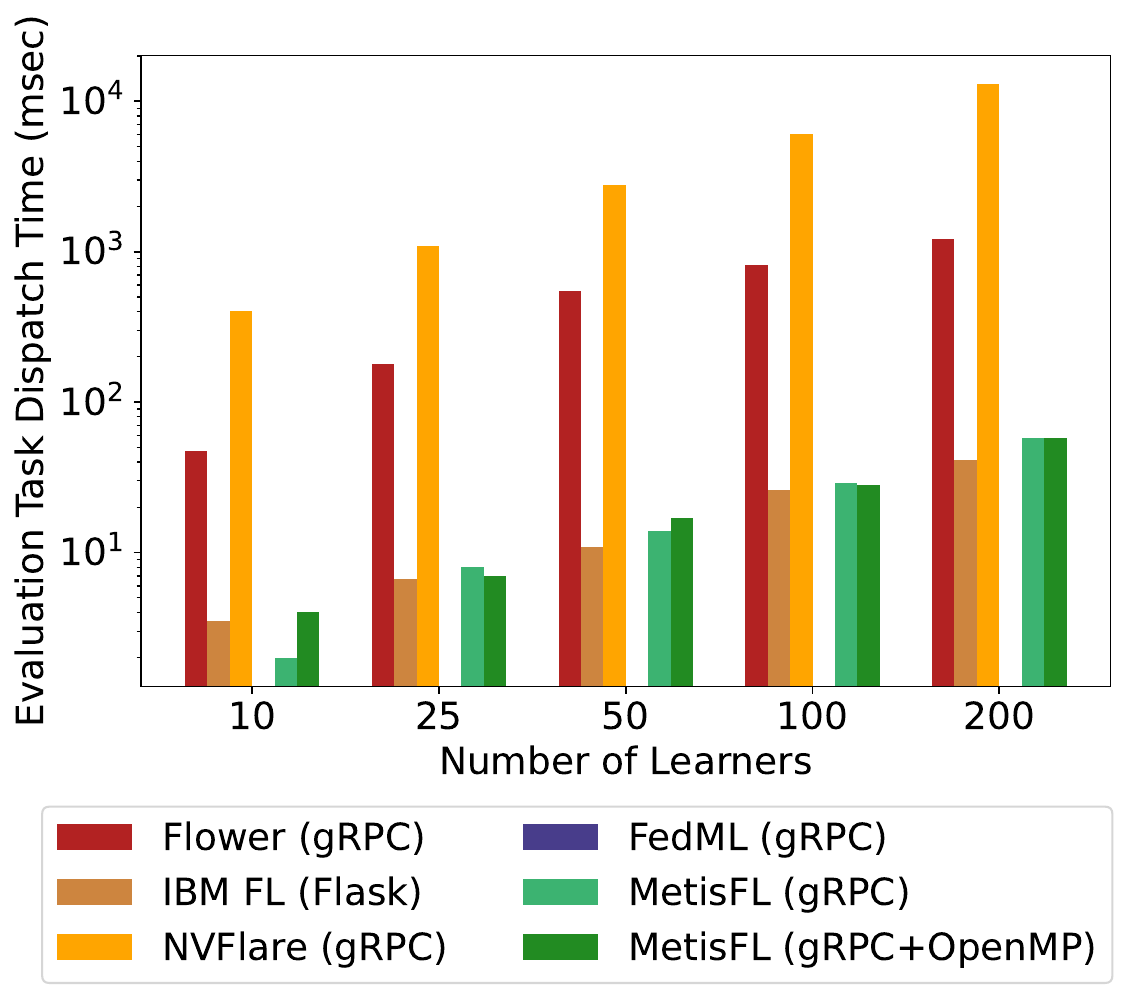}
        \label{fig:params_1M_eval_dispatch_time}
    }
    \subfloat[Eval Round]{
        \centering
        \includegraphics[width=0.25\linewidth]{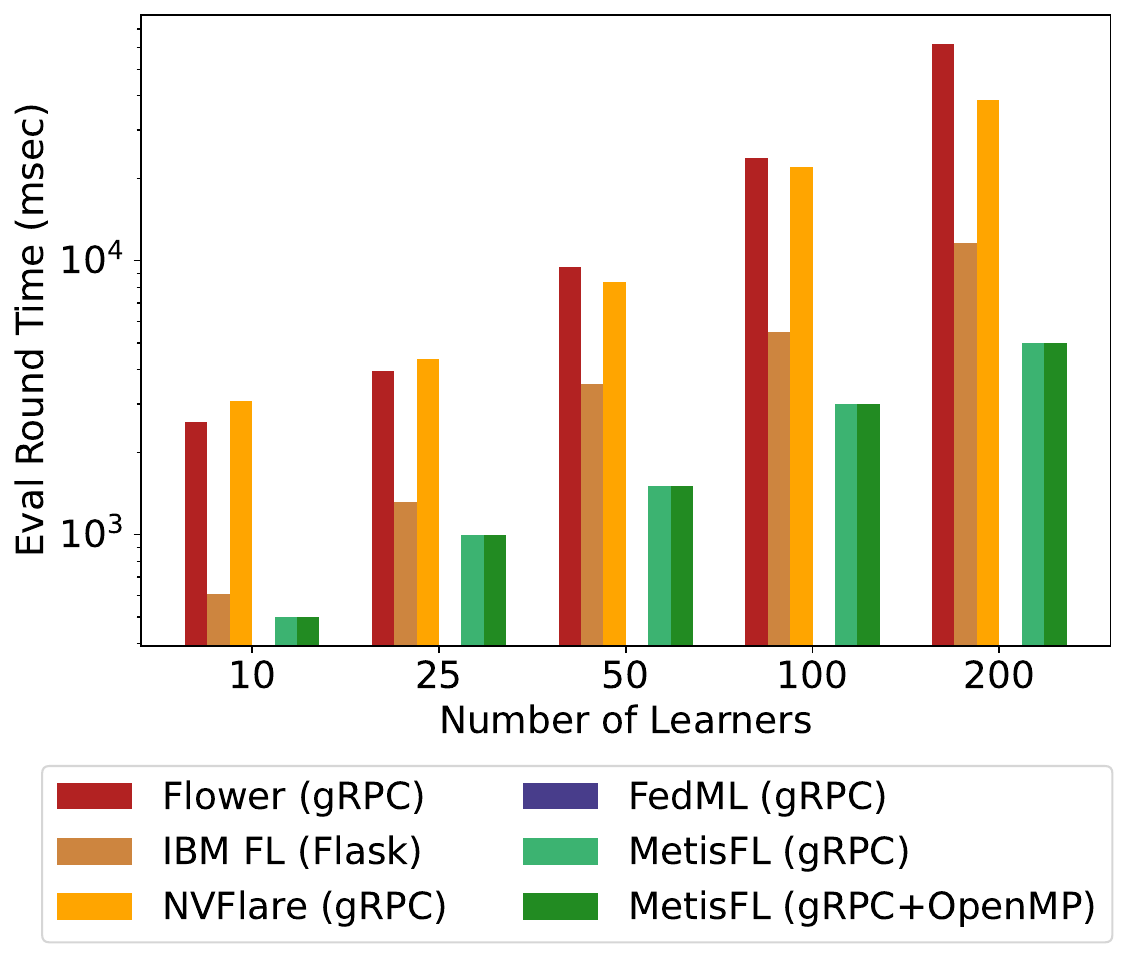}
        \label{fig:params_1M_eval_round_time}
    }
    \subfloat[Federation Round]{
        \centering
        \includegraphics[width=0.25\linewidth]{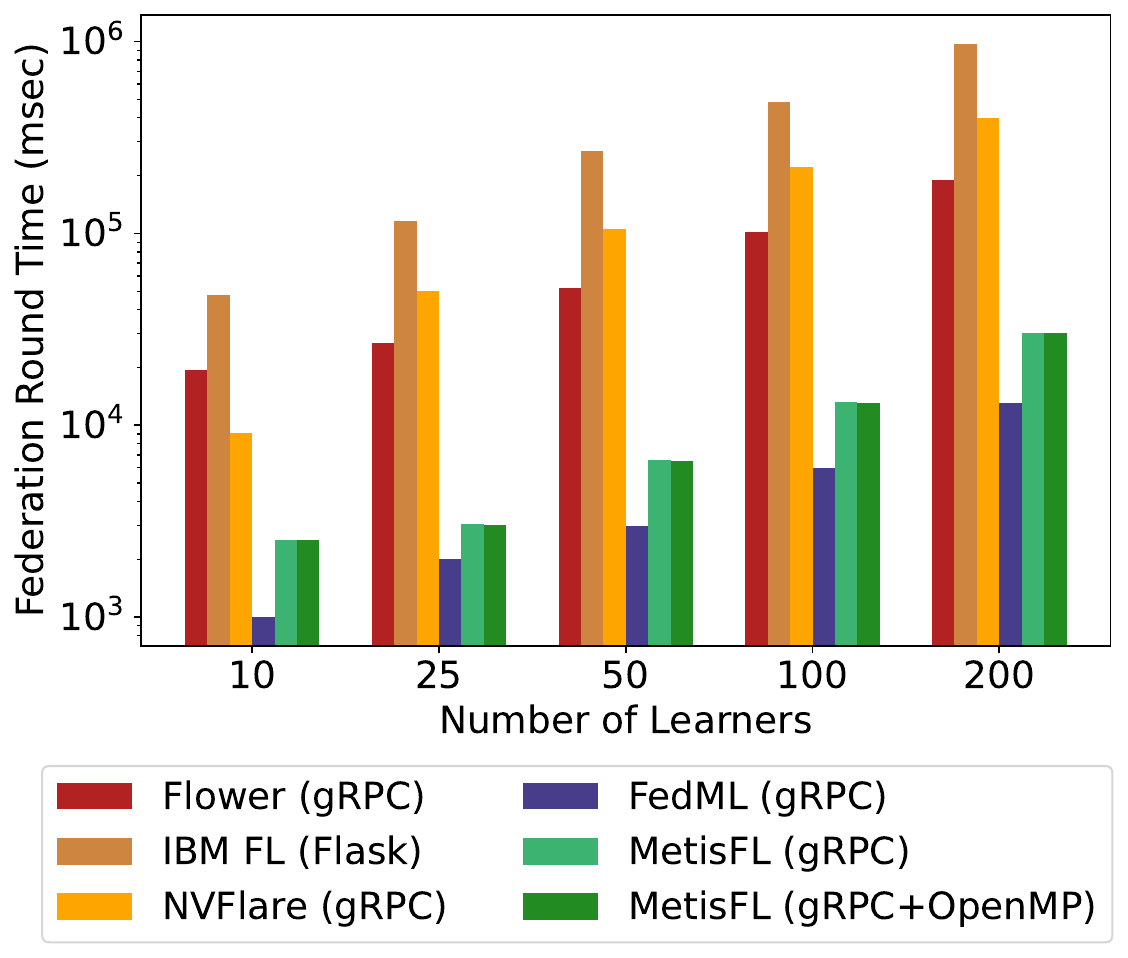}
        \label{fig:params_1M_federation_round_time}
    }    

    \captionsetup{justification=centering}
    \caption{FL frameworks operations comparison for 1M parameters (the y-axis is in logscale).}
    \label{fig:FL_Systems_Quantitative_Comparison_1M}
\end{figure*}

\begin{figure*}

    \centering
    \subfloat[Train Task Dispatch]{
        \centering
        \includegraphics[width=0.25\linewidth]{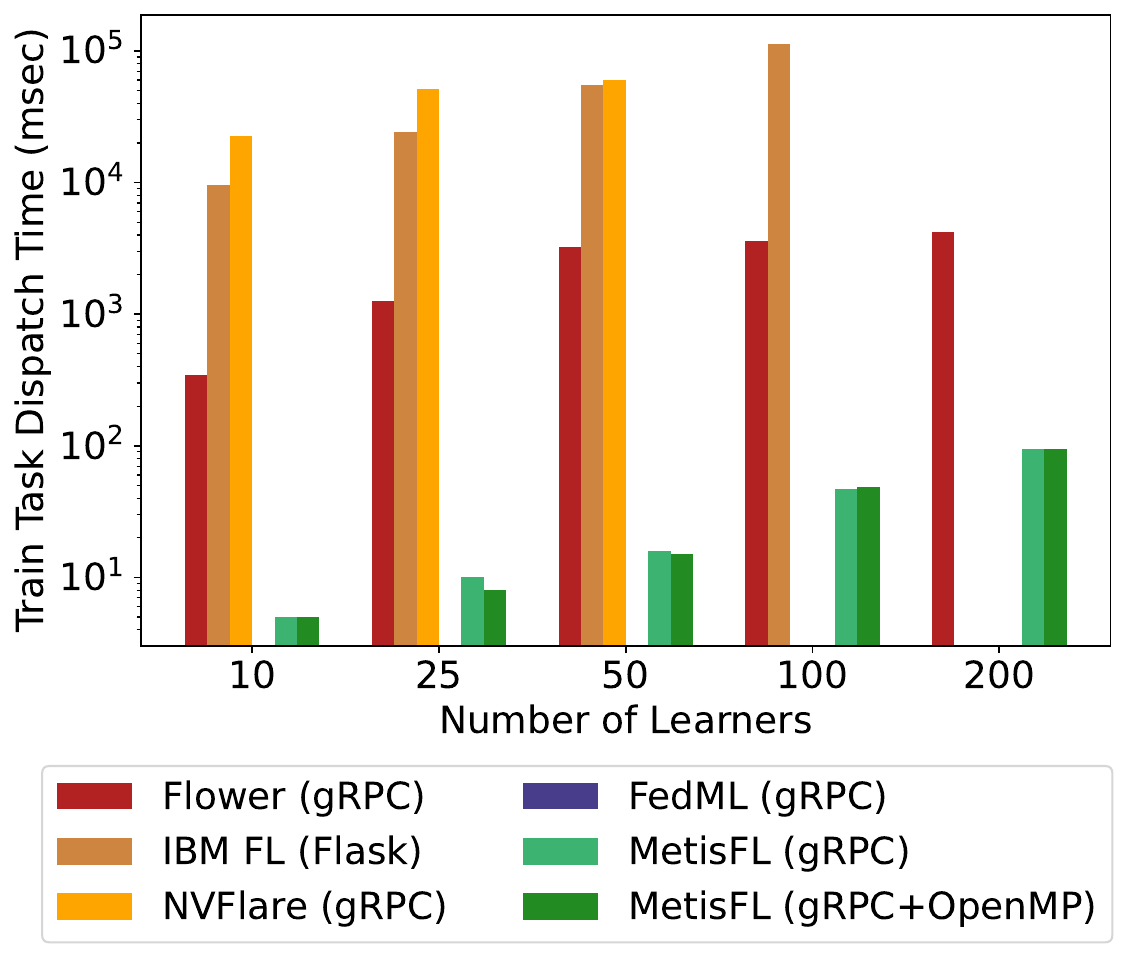}
        \label{fig:params_10M_train_dispatch_time}
    }
    \subfloat[Train Round]{
        \centering
        \includegraphics[width=0.25\linewidth]{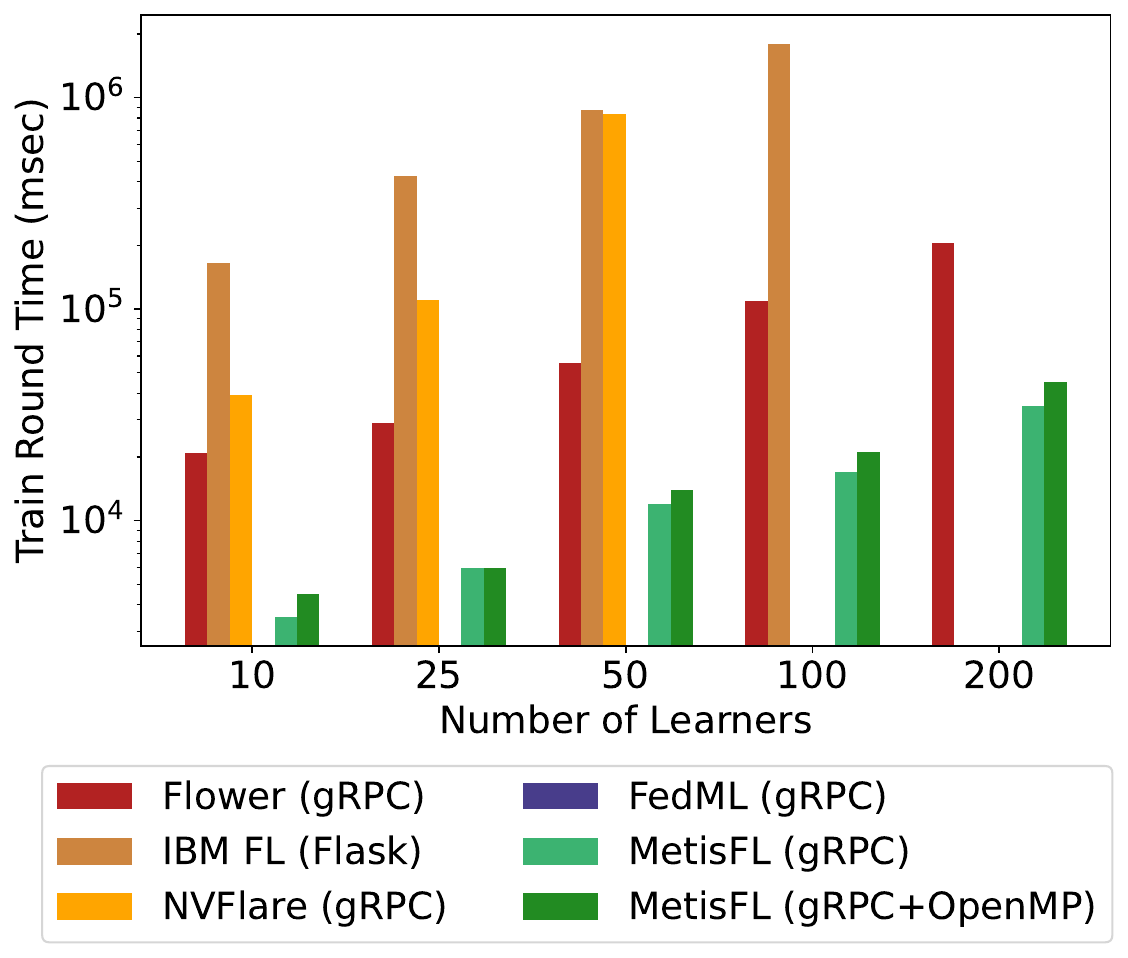}
        \label{fig:params_10M_train_round_time}
    }    
    \subfloat[Aggregation Time]{
        \centering
        \includegraphics[width=0.25\linewidth]{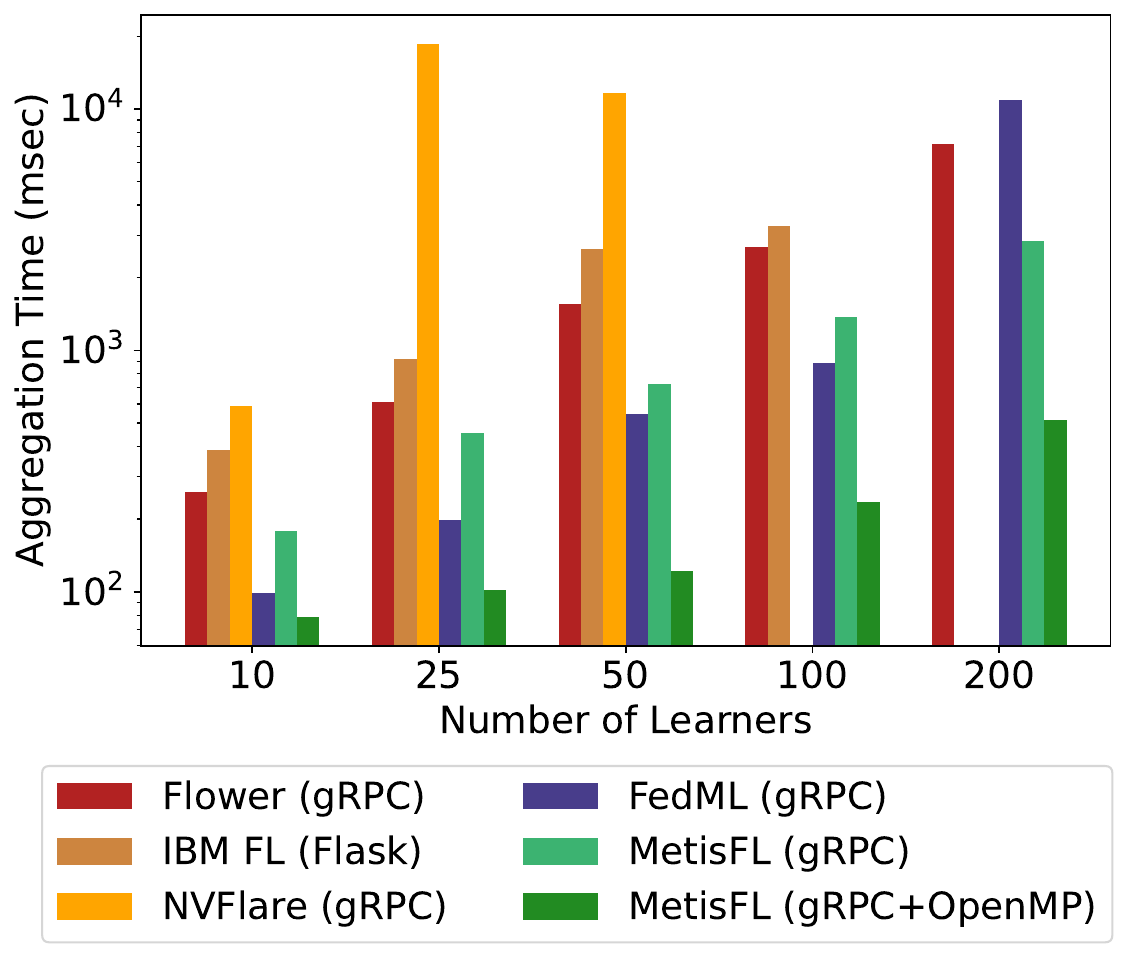}
        \label{fig:params_10M_agg_time}
    }

    \subfloat[Eval Task Dispatch]{
        \centering
        \includegraphics[width=0.25\linewidth]{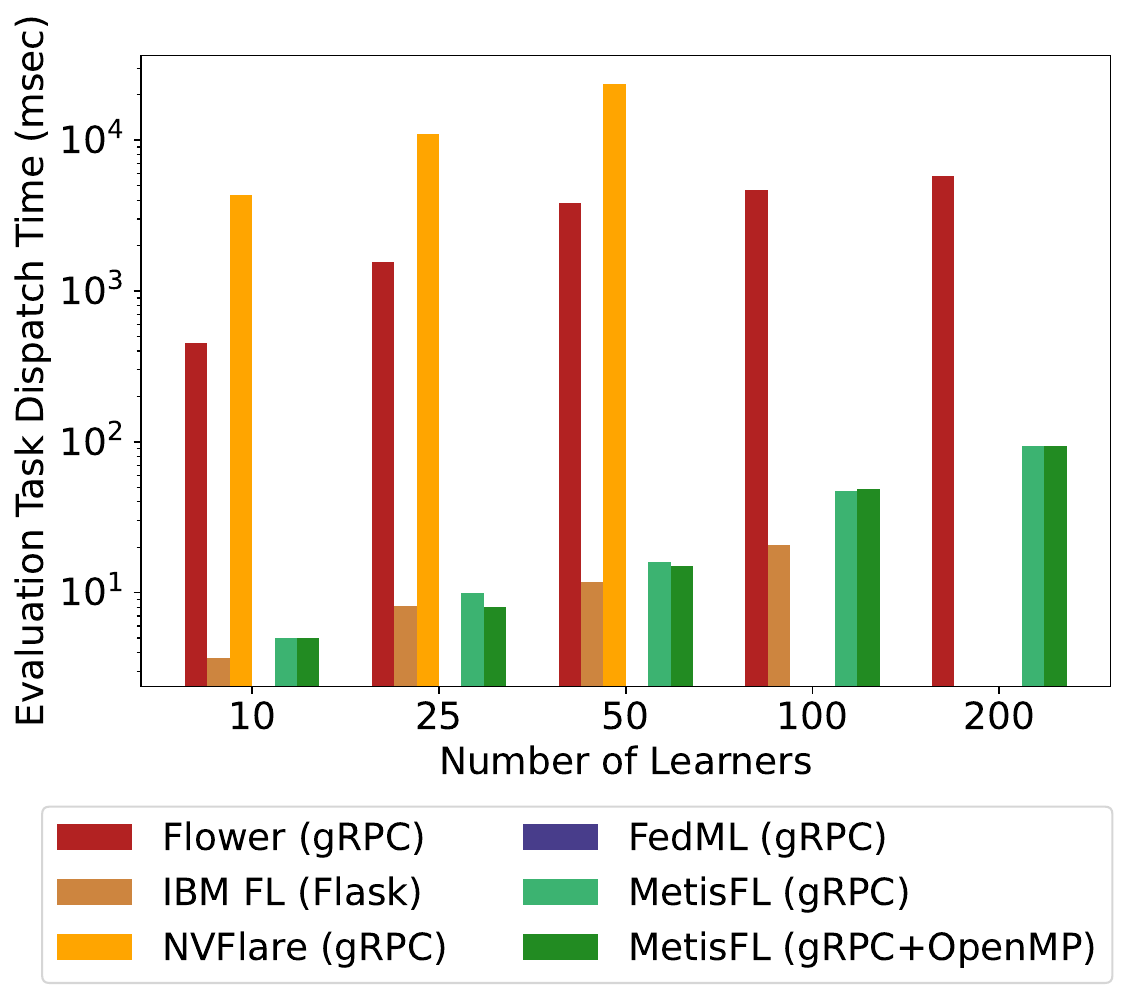}
        \label{fig:params_10M_eval_dispatch_time}
    }
    \subfloat[Eval Round]{
        \centering
        \includegraphics[width=0.25\linewidth]{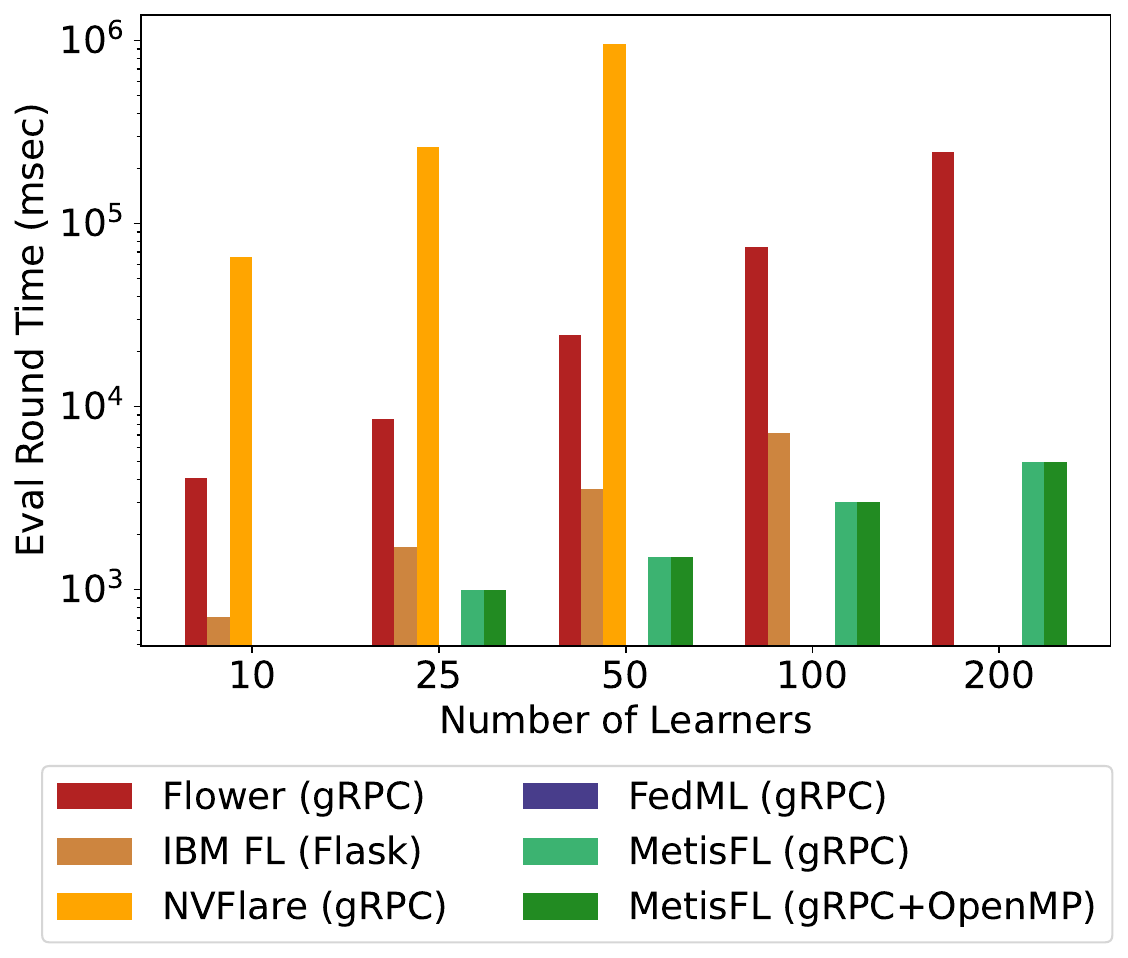}
        \label{fig:params_10M_eval_round_time}
    }
    \subfloat[Federation Round]{
        \centering
        \includegraphics[width=0.25\linewidth]{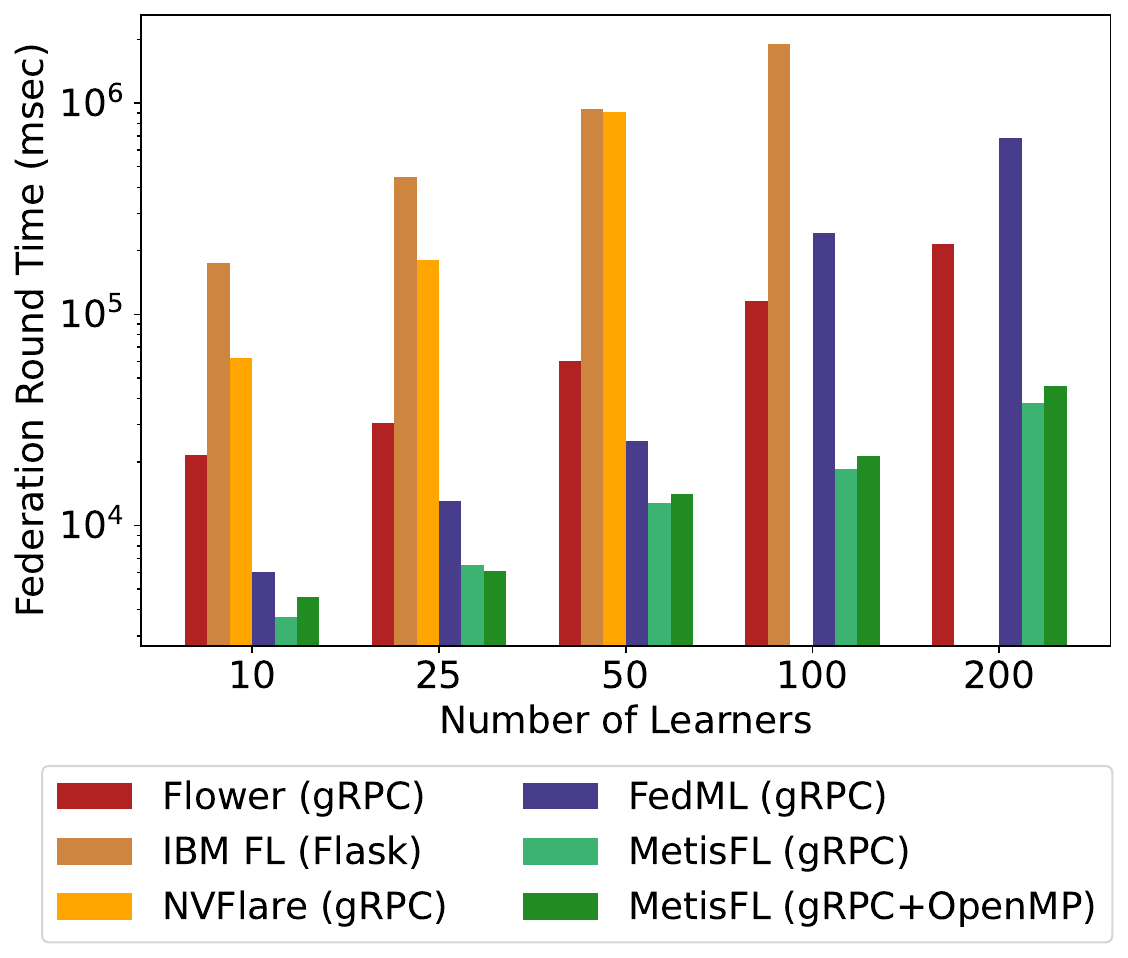}
        \label{fig:params_10M_federation_round_time}
    }    

    \captionsetup{justification=centering}
    \caption{FL frameworks operations comparison for 10M parameters (the y-axis is in logscale).}
    \label{fig:FL_Systems_Quantitative_Comparison_10M}    
    
\end{figure*}

For the quantitative evaluation, presented in Figures~\ref{fig:FL_Systems_Quantitative_Comparison_100k},~\ref{fig:FL_Systems_Quantitative_Comparison_1M}, and~\ref{fig:FL_Systems_Quantitative_Comparison_10M}, we conduct end-to-end system stress tests across all frameworks over an increasing number of participating learners: \{10, 25, 50, 100, 200\}, and model sizes: \{100k, 1M, 10M\} parameters. We test all frameworks in a synchronous FL setting, using FedAvg as the global aggregation rule and with all learners participating at every federation round. To generate the different model sizes, we define an MLP architecture with 100 densely connected (hidden) layers and a constant number of parameters per layer~\footnote{100k: 32 params/layer, 1M: 100params/layer, 10M: 320params/layer}. For training and evaluating the model, we use the HousingMLP dataset~\cite{breiman2017classification} to generate the train and test datasets. Given that our analysis aims to stress test the framework's efficiency and not the FL model's learning performance, we randomly pick (with replacement) and assign 100 samples per learner for model training and testing. As local model optimizer, we use Vanilla SGD, and the batch size is set to 100 (both for training and testing). 

All experiments were conducted in a simulated federated environment on the same host machine equipped with 32cores Intel(R) Xeon(R) Gold 5217 CPU @ 3.00GHz, 251GB memory, and 8TB disk space. All conducted experiments were CPU-intensive, with each learner's workload being CPU-bound, and no GPU was employed during federated training. With regard to frameworks' versioning, we used NVFlare 2.3.1, Flower 1.2.0, FedML 0.7.6, and IBM FL 1.1.0. Our evaluation does not report OpenFL results because of the framework's complexity in incorporating new models and logging features~\footnote{We will try to include them in a future version of our work.}. For all frameworks, we used gRPC as the services' communication layer, except for IBM FL, where we used the FLASK API. For Flower, IBM FL, and MetisFL, we used Keras as the backend NN engine, and for NVFlare and FedML, we used PyTorch. 

With respect to the frameworks' measurement reporting, in the Figures~\ref{fig:FL_Systems_Quantitative_Comparison_100k},~\ref{fig:FL_Systems_Quantitative_Comparison_1M}, and~\ref{fig:FL_Systems_Quantitative_Comparison_10M}, for all other frameworks, except for FedML, we were able to capture all required metrics. For FedML, in particular, we only report the aggregation and federation time wall-clock time because it was hard to navigate into the codebase and find the required code segments where the logging functionality for the rest of the operations must be placed. Moreover, concerning execution failures, NVFlare and IBM FL did not run in the federated environment of 10M parameters for 100 and 200 learners and 200 learners, respectively.

In the case of MetisFL, we ran each MetisFL for each environment twice, one with the OpenMP enabled (MetisFL gRPC + OpenMP) and one without (MetisFL gRPC). As expected, OpenMP model aggregation is 10 times faster than no parallelization (cf. MetisFL w/ and w/out parallelization in Figures~\ref{fig:params_100k_agg_time},~\ref{fig:params_1M_agg_time},~\ref{fig:params_10M_agg_time}) and almost 100 times faster than other frameworks, especially in the case of federation environments with large models (cf. MetisFL w/ OpenMP to other FL frameworks in Figure~\ref{fig:params_10M_agg_time}). Since parallelization is only used during model aggregation, all other operations performed by MetisFL with and without OpenMP have similar and almost identical performances across all environments. 

When comparing each federated operation in isolation, we observe that most systems' overhead is attributed to the dispatch time of training and evaluation tasks (cf. Figures~\ref{fig:params_100k_train_dispatch_time},~\ref{fig:params_100k_eval_dispatch_time},~\ref{fig:params_1M_train_dispatch_time},~\ref{fig:params_1M_eval_dispatch_time},~\ref{fig:params_10M_train_dispatch_time},~\ref{fig:params_100k_eval_dispatch_time}). MetiFL has a relatively smaller overhead for these operations than other systems because the MetisFL controller submits the tasks to the learners through asynchronous callbacks, and all model tensors are transmitted as a protobuf bytes data type. 

From our evaluation, we also observe that all NVFlare's operations, especially its training and evaluation dispatch tasks times, underperform the corresponding operations from all other frameworks. When it comes to IBM FL, even though IBM FL does great work on performing extremely fast evaluation task dispatching and round execution, it underperforms other systems when executing the corresponding training operations. However, across all operations and frameworks, Flower has the most consistent behavior. As the federation size is doubled or the federation model becomes larger, Flower's execution latency is linearly increasing.

Finally, when comparing all frameworks holistically with respect to the federation round execution time (cf. Figures~\ref{fig:params_100k_federation_round_time},~\ref{fig:params_1M_federation_round_time},~\ref{fig:params_10M_federation_round_time}), we observe that as the model sizes become larger and the federation size increases, the impact becomes much more significant. For instance, even though FedML is outperforming all frameworks in the federation environments with model sizes of 100k and 1M parameters, its performance starts to deteriorate when the size of the federated model becomes larger, i.e., 10M parameters. However, when comparing MetisFL to FedML in these challenging environments, we can observe that MetisFL leads to a 10-fold improvement. In general, all frameworks seem to underperform in these computationally demanding environments, with the overall federation execution time increasing almost non-linearly as the size of the federation doubles (cf. Table~\ref{tbl:FedRoundTime_10M}), e.g., 13.85s $\rightarrow$ 25.31s $\rightarrow$ 243.54 for FedML and 448.57s $\rightarrow$ 934.73s $\rightarrow$ 1,914.92s for IBM FL, while Flower's overall federation execution time is doubled every time the number of participating learners is doubled, i.e., 60.44 $\rightarrow$ 116.05 $\rightarrow$ 216.38.

\begin{table}[htbp]
\centering
\begin{tabular}{@{}cccccc@{}}
\toprule
\#Learners & NVFlare & Flower & FedML & IBM FL & MetisFL \\

\cmidrule(lr){1-1}
\cmidrule(lr){2-2}
\cmidrule(lr){3-3}
\cmidrule(lr){4-4}
\cmidrule(lr){5-5}
\cmidrule(lr){6-6}

10 & 62.32 & 21.51 & 6.21 & 175.19 & 4.58 \\
25 & 180.63 & 30.71 & 13.85 & 448.57 & 6.10 \\
50 & 907.92 & 60.44 & 25.31 & 934.73 & 14.13 \\
100 & N/A & 116.05 & 243.54 & 1,914.92 & 21.28 \\
200 & N/A & 216.38 & 683.21 & N/A & 45.61 \\

\bottomrule
\end{tabular}
\captionsetup{justification=centering}
\caption{Federation Round Time (secs) for 10M Parameters.}
\label{tbl:FedRoundTime_10M}
\end{table}

\section{Discussion}\label{sec:Discussion}
We presented the Metis Federated Learning framework, a novel system for accelerating the training of large-scale FL workflows. By re-engineering the federation controller and treating it as the system's first-class citizen, MetisFL leads to a powerful 10-fold and 100-fold wall-clock time reduction compared to existing leading FL frameworks. In our immediate future work, we plan to further enhance the computational capabilities of the controller by studying large-scale FL workflows in environments consisting of thousands and millions of learners and extremely large models (+100M parameters) in both synchronous and asynchronous settings. Given the amount of computational resources required in these environments, the controller may not be able to store all required local models in memory. In such settings, we plan to incorporate different model stores (e.g., distributed key-value or on-disk model stores) and understand their trade-offs in FL workflows' convergence. We also plan to perform a more systematic performance analysis with respect to the cryptographic schemes employed by each framework and realize how the various FL workflows are affected.

\section*{Acknowledgements}
This research was supported in part by the Defense Advanced Research Projects Agency (DARPA) under contract HR0011\-2090104, and in part by the National Institutes of Health (NIH) under grants U01AG068057 and RF1AG051710.  The views and conclusions contained herein are those of the authors and should not be interpreted as necessarily representing the official policies or endorsements, either expressed or implied, of DARPA, NIH, or the U.S. Government.

\bibliography{main}
\bibliographystyle{tmlr}

\appendix

\section{FL Frameworks Configuration}

\noindent\paragraph{\textbf{Flower}} Flower is a widely used open-source federated learning framework. Flower’s architecture addresses critical distributed computing challenges such as scalability, client and communication heterogeneity, and system flexibility by intelligently organizing workload infrastructure between the server and client. To perform the Flower experiments, we leveraged the system’s federated averaging strategy abstraction to represent the coordinating server to the distributed client set. To set up all experimental environments, we created a Client Learner class, which defined each learner's training and evaluation behavior in the federation. Then, we defined the client-factory function, which creates the required number of learners for the simulation and initializes each client with the model training and test data. We used system-native logging to record timestamp differences between specific code segments to benchmark performance successfully. We added timestamps wherever appropriate to record each experiment's aggregation, dispatch, and overall wall-clock time.

\noindent\paragraph{\textbf{FedML}} To configure the environments for our experiments, we installed FedML using pip inside a docker container. We changed the code in the Python source files installed using pip to record timestamps for the model aggregation time, federation, training, and evaluation round times, along with the associated dispatch time for the training and test tasks. To define the simulation for our use cases, we defined a data loader class for reading the input training data to our model. We registered the model by making code changes to the function Model Hub. To stress-test the system with different communication protocols, we run the system with two different back-end communication services: gRPC and MQTT.

\noindent\paragraph{\textbf{IBM FL}} We spawn the IBM FL framework in a containerized environment. Specific terminology within the framework designates the Aggregator as the coordinating server and the distributed clients as Parties. Arrangement of both the aggregator and parties is handled by IBM FL native python scripts that process experiment parameters to create configuration files for the aggregator and parties. The federated learning training algorithm is assigned to the aggregator defined as a fusion handler. In our experiments, we selected the native FedAvgFusionHandler. Each party is assigned our custom Housing MLP model specified with parameter size at configuration time. To execute our experiments, we spawned one process solely responsible for the aggregator and a variable number of party processes set to the number of parties. We added timestamps in the code logic surrounding the global model aggregation function, the functions that execute the dispatch and evaluate logic. Moreover, to enable easier initialization of the federated environments, we had to tailor the initialization logic of the Party class code so that it could execute model training as soon as the aggregator directed it. By doing so, we did not have to initialize the Party class through command line instructions, which was extremely cumbersome when considering multiple learners for each experiment.

\noindent\paragraph{\textbf{Nvidia Flare}} NVFlare operates as a generic open-source software development kit (SDK) to facilitate distributed collaborative computing in multithreaded environments. Central to this framework’s architecture is a flexible implementation of communication protocols carried out by two distinct entities: Controllers and Workers. In our FL experiments, we utilized the functionality of these communication protocols to represent the principal roles of the solitary server (Controller) and the set of distributed clients (Workers). An example of such a communication protocol used in tandem with Federated Averaging is the “Broadcast and Wait” protocol. In this protocol, the server broadcasts a task and global information to the clients. Each client then locally executes the task and sends the client’s local updates to the orchestrating server for aggregation. The framework's infrastructure places a strong emphasis on workflows. Namely, we used the Scatter and Gather as well as Global Model Evaluation workflows. To execute the experiments, NVFLARE provides a robust simulator command line interface, which we employ to configure various experiment details, including the client count, number of threads, and machine learning tasks. We utilized developer-friendly built-in logging methods to record training and evaluation times and capture system-related metrics. By harnessing the robust foundational system, effective communication protocol, user-friendly simulator, and reliable logging, we successfully benchmarked the approach with metrics pertinent to this paper.

\section{MetisFL Internal Procedures}
In this section, we describe in more detail some of the internal mechanisms of MetisFL execution flow. Specifically, we consider a synchronous FL setting, where federated training and evaluation occur over a series of federation rounds, with every federation round consisting of a federated training round followed by a federated evaluation round.

\begin{figure}[htpb]
  \centering  
  \includegraphics[scale=0.5]
  {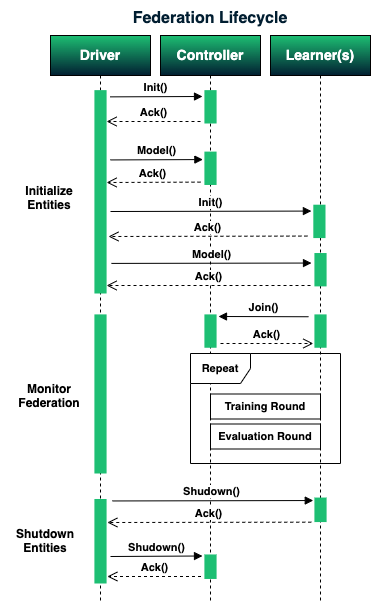}
  \caption{MetisFL Federation Round Flow.}
  \label{fig:MetisFL_Federation_Round_Flow}
\end{figure}

\noindent\paragraph{\textbf{Federation Round Flow}}
The Federation Round lifecycle (see Figure~\ref{fig:MetisFL_Federation_Round_Flow}) consists of three core steps: \textit{Initialization, Monitoring, Shutdown}. The driver initializes the controller process at the remote host (or localhost) and receives an acknowledgment when the remote process is alive. After that, the driver sends the initial state of the model (just the model tensors, not the actual architecture) to the controller and proceeds with learners' initialization. Once all learners are initialized, the driver sends the model (model tensors \& architecture) to every other learner. The driver ships the actual model architecture to the learners because they are required to perform training and evaluation on their local private datasets. In contrast, the controller is only responsible for orchestrating the federation workflow and aggregating learners' weights.

Once learners register with the controller (join the federation), the training and evaluation of the federated model occur for multiple federation rounds. Within every federation round, the federated model is sent for training and, subsequently, for evaluation to all participating learners. At this point, the driver monitors the lifecycle of the federation and periodically pings (heartbeat) remote processes. Once any of the federated training termination criteria is met, such as the execution wall-clock time or a number of federation rounds, then the driver sends a shutdown signal to all processes, first to the learners and then to the controller.

\begin{figure}[htpb]
  \centering  
  \includegraphics[scale=0.5]
  {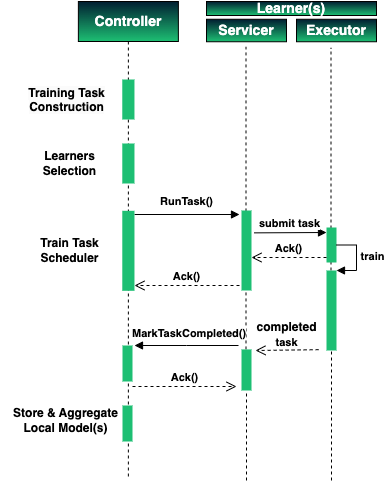}
  \caption{MetisFL Training Round Flow.}
  \label{fig:MetisFL_Training_Round_Flow}
\end{figure}

\noindent\paragraph{\textbf{Training Round Flow}} Figure~\ref{fig:MetisFL_Training_Round_Flow} presents how the federated training round is executed within MetisFL. Before the training round starts, the controller creates/defines the model training task and selects the learners participating in model training. Once the learners have been selected the train task scheduler sends the training task to every participating learner (RunTask request).

The Learner(s) entity receives the task through the Learner Servicer process and submits the training task to the training task pool executor running in the background. Upon task submission, the executor replies with an Acknowledgment (Ack) message that the servicer relays to the controller. Note that the status of the Ack message is false when the training task is not submitted or received or any unexpected failure occurs.

The submitted training task is registered with a callback function that will handle the completed training task when it is completed. Once this occurs, the servicer sends a MarkTaskCompleted request to the controller containing the learner's local model and other execution metadata related to model training (e.g., training time per batch, number of completed steps, and epochs). Finally, the controller stores and aggregates all received local models.

To improve the scheduling of training tasks, the controller submits the tasks to the learners as asynchronous calls, meaning that the controller does not wait for the training task to complete. In other words, the controller submits the task, but the learner needs to inform the controller when its local training is complete.

\begin{figure}[htpb]
  \centering  
  \includegraphics[scale=0.5]
  {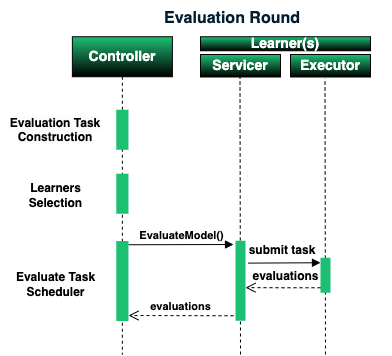}
  \caption{MetisFL Evaluation Round Flow.}
  \label{fig:MetisFL_Evaluation_Round_Flow}
\end{figure}

\noindent
\paragraph{\textbf{Evaluation Round Flow}} In Figure~\ref{fig:MetisFL_Evaluation_Round_Flow} we present the execution of the federated evaluation round. Similar to the training round, the evaluation round starts with the controller constructing the evaluation task and selecting the learners participating in the evaluation of the global model. Once these steps are defined, the evaluation task scheduler sends an EvaluateModel request to all participating learners and receives the respective model evaluations. Compared to the training tasks, the evaluation tasks are synchronous calls, meaning that the controller keeps the connection alive till the evaluation of the model is complete.

\begin{figure}[htpb]
  \centering  
  \includegraphics[width=0.5\linewidth]{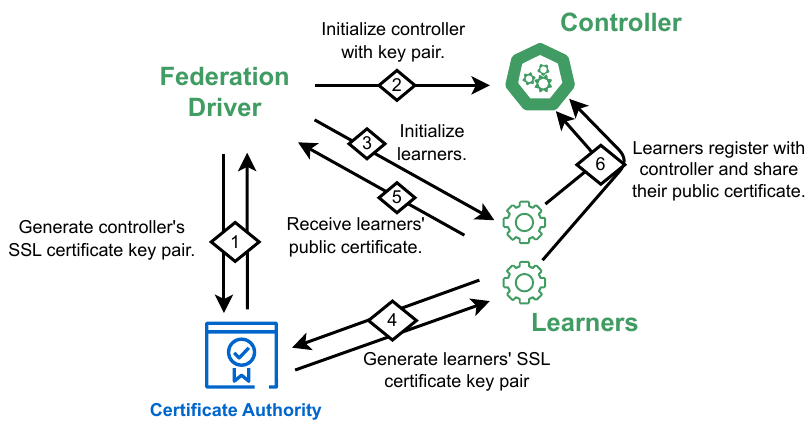}
  \caption{MetisFL with SSL Execution.}
  \label{fig:MetisFL_SSL}
\end{figure}

\noindent\paragraph{\textbf{Secure Sockets Layer}} Figure~\ref{fig:MetisFL_SSL} shows how SSL is enabled within MetisFL. From the perspective of SSL connectivity, in a FL setting every participating entity (controller, learner) acts both as a client and as a server at different points of the federated execution workflow. Therefore, when establishing SSL-enabled connections we need first to generate the pair of files (private key, public certificate) that will be used by the server process running at a given host (e.g., learner/controller). This will allow the gRPC server to receive secure connections from requesting clients. When the controller receives model update requests from the learners, the controller acts as a server and the learners as the server's clients. Similarly, when a learner receives local model training requests from the controller, the learner acts as a server and the controller as a client.

In a MetisFL simulated environment, the internal mechanism for enabling SSL connectivity starts with the federation driver. The driver uses the defined SSL certificates (self-signed or trusted authority) to start the federation controller and spawn its server process with the provided key pair. Analogously, the driver initializes the learner processes using the defined pair. Finally, during the learner-controller registration, the learners share their public certificate with the controller as part of the exchanged message.

In a production environment, though, it is not required for the SSL certificates to be generated by the driver. The certificates can be developed independently by each process (controller, learner), and then the public certificates can be shared with the driver to establish secure connections wherever needed.

\end{document}